\documentclass{article}
\usepackage{iclr2027_conference,times}

\usepackage{amsmath,amsfonts,bm}

\def\eqref#1{equation~\ref{#1}}

\def\1{\bm{1}}

\DeclareMathAlphabet{\mathsfit}{\encodingdefault}{\sfdefault}{m}{sl}
\SetMathAlphabet{\mathsfit}{bold}{\encodingdefault}{\sfdefault}{bx}{n}

\usepackage[table]{xcolor}
\usepackage{graphicx}
\usepackage{tikz}
\usetikzlibrary{arrows.meta,calc,positioning}
\usepackage{amssymb}
\usepackage{booktabs}
\usepackage{multirow}
\usepackage{array}
\usepackage{makecell}
\usepackage{longtable}
\usepackage{subcaption}
\usepackage{algorithm}
\usepackage{algpseudocode}
\usepackage[T1,OT1]{fontenc}
\usepackage{listings}
\usepackage{enumitem}
\usepackage{hyperref}
\usepackage[section]{placeins}
\usepackage{float}
\usepackage{url}
\usepackage[capitalize]{cleveref}

\DeclareRobustCommand{\ours}{\textsc{Cadet}}
\DeclareRobustCommand{\cadetdata}{RouteDemos}
\setlist[itemize,enumerate]{leftmargin=*,itemsep=1pt,parsep=0pt}
\DeclareRobustCommand{\pcbworld}{PCBWorld}

\DeclareRobustCommand{\kicad}{\textsc{KiCad}}
\renewcommand{\eqref}[1]{\textup{(\ref{#1})}}

\newcommand{\lognormal}{\mathrm{LN}}
\newcommand{\rev}[1]{#1}

\newenvironment{narrowalgorithm}[2][!t]
  {\begin{figure}[#1]\centering\setlength{\intextsep}{4pt}\begin{minipage}{#2}\begin{algorithm}[H]}
  {\end{algorithm}\end{minipage}\end{figure}}

\crefname{figure}{Figure}{Figures}
\crefname{subfigure}{Figure}{Figures}

\makeatletter
\renewcommand{\theHALG@line}{\thealgorithm.\arabic{ALG@line}}
\makeatother
\newcommand{\promptfont}{\fontencoding{T1}\renewcommand{\ttdefault}{zi4}\ttfamily}
\lstdefinestyle{prompt}{
  basicstyle=\promptfont\footnotesize,
  columns=fullflexible,
  keepspaces=true,
  breaklines=true,
  breakindent=1.5em,
  frame=single,
  framerule=0.5pt,
  rulecolor=\color{black!50},
  backgroundcolor=\color{black!4},
  framesep=6pt,
  xleftmargin=8pt,
  aboveskip=4pt,
  belowskip=4pt,
  extendedchars=true,
  upquote=true,
  inputencoding=utf8,
  moredelim=**[s][\color{teal!70!black}]{\{}{\}},
  literate={—}{{---}}1 {≥}{{$\geq$}}1,
}

\definecolor{cdtInk}{HTML}{243746}
\definecolor{cdtBlue}{HTML}{356B9A}
\definecolor{cdtTeal}{HTML}{168577}
\definecolor{cdtAmber}{HTML}{B8791C}
\definecolor{cdtRed}{HTML}{B7504D}
\definecolor{cdtGray}{HTML}{89949C}
\definecolor{cdtPale}{HTML}{F3F6F7}
\tikzset{
  cdt text/.style={font=\sffamily\fontsize{8}{9.4}\selectfont,text=cdtInk,align=center},
  cdt heading/.style={cdt text,font=\sffamily\bfseries\fontsize{8}{9.4}\selectfont},
  cdt small/.style={cdt text,font=\sffamily\fontsize{7}{8.2}\selectfont},
  cdt box/.style={draw=cdtGray!60,fill=white,rounded corners=2pt,line width=.5pt},
  cdt arrow/.style={-{Stealth[length=1.6mm,width=1.2mm]},draw=cdtInk!80,line width=.7pt},
  cdt retry/.style={cdt arrow,draw=cdtRed,dashed},
  cdt trace/.style={draw=cdtTeal,line width=1.4pt,rounded corners=1.3pt},
  cdt pad/.style={draw=cdtTeal,fill=white,line width=1pt},
  cdt proposal/.style={draw=cdtAmber,fill=cdtAmber!12,dashed,line width=.8pt},
  cdt board/.style={draw=cdtGray!75,fill=cdtPale,rounded corners=2pt,line width=.6pt},
}

\newcommand{\cadetBoard}[1]{%
  \draw[cdt board] (0,0) rectangle (2.5,1.35);
  \filldraw[fill=cdtGray!18,draw=cdtGray!45,rounded corners=1pt]
    (1.05,.45) rectangle (1.5,.95);
  \ifnum#1>0
    \draw[cdt trace] (.35,.3) -- (.8,.3) -- (.8,1.05);
  \fi
  \ifnum#1=3
    \draw[cdt trace] (.8,1.05) -- (1.72,1.05) -- (2.12,.65);
  \fi
  \draw[cdt pad] (.35,.3) circle (.09);
  \draw[cdt pad] (.8,1.05) circle (.09);
  \ifnum#1=2
    \draw[cdt proposal] (2.12,.65) circle (.12);
    \draw[draw=cdtAmber,densely dotted,line width=.7pt]
      (.8,1.05) -- (2.12,.65);
  \fi
  \ifnum#1=3 \draw[cdt pad] (2.12,.65) circle (.09); \fi
  \ifnum#1=0
    \draw[cdt pad] (2.12,.65) circle (.09);
    \draw[draw=cdtTeal!55,densely dashed,line width=.55pt]
      (.35,.3) -- (.8,1.05) -- (2.12,.65);
  \fi
}

\title{CADET: Construction of PCB Routing\\Instances with Executable Demonstrations}
\author{%
\begin{tabular*}{\dimexpr\textwidth-2\tabcolsep\relax}[t]{@{\extracolsep{\fill}}ccc@{}}
Hyungseok Song & Won-Seok Choi & Seohui Bae\\
{\normalfont\small hyungseok.song@lgresearch.ai} & {\normalfont\small wonseok.choi@lgresearch.ai} & {\normalfont\small seohui.bae@lgresearch.ai}\\
{\normalfont\small LG AI Research} & {\normalfont\small LG AI Research} & {\normalfont\small LG AI Research}\\[6pt]
Han-Seul Jeong & Junseok Park & Youngjoon Park\\
{\normalfont\small hanseul.jeong@lgresearch.ai} & {\normalfont\small frank.park227@lgresearch.ai} & {\normalfont\small yj.park@lgresearch.ai}\\
{\normalfont\small LG AI Research} & {\normalfont\small LG AI Research} & {\normalfont\small LG AI Research}\\[6pt]
 & Soonyoung Lee & \\
 & {\normalfont\small soonyoung.lee@lgresearch.ai} & \\
 & {\normalfont\small LG AI Research} & \\
\end{tabular*}}

\iclrfinalcopy

\begin{document}
\maketitle
\lhead{Preprint.}

\begin{abstract}
Printed circuit board (PCB) routing is the stage of physical layout that finds paths connecting the placed pads of each net. Public datasets in this domain are few, and most contain only the final states of routed boards, not demonstrations of the electronic design automation (EDA) engine actions that produce them. To close this gap, we propose \ours{} (Constructing executAble Demonstrations for Engine-grounded Training), which jointly generates synthetic boards and demonstrations built from EDA engine actions. The resulting \cadetdata{} provides 100,000 synthetic boards routed with zero design rule violations (DRVs), together with their demonstrations. Their distributions over 17 statistics resemble those of 142 real reference boards. \ours{} trains a routing policy on these demonstrations with behavior cloning and fine-tunes it with reinforcement learning (RL). On real test boards, the policy raised the rate of complete DRV-free routing from \rev{38.0\% to 67.2\%} over an RL-only baseline. \rev{This rate was comparable to that of the open-source autorouter FreeRouting and exceeded those of the other baselines, including GPT-6 Astra.} We release the dataset and its generation code as open source at {\urlstyle{same}\url{https://github.com/LGAI-Research/CADET}}.
\end{abstract}

\section{Introduction}
\label{sec:intro}

Printed circuit board (PCB) routing is the stage of physical layout that electrically connects the pads of each net under design rules. As pin counts and densities grow, manual routing becomes increasingly time-consuming, and its automation grows in importance~\citep{unetastar2023}. Rule-based autorouters still dominate in practice~\citep{freerouting,kicadroutingtools}. Most learning-based routers either route cell by cell on a grid~\citep{liao2020globalrouting,jumanji} or delegate path construction to an existing router~\citep{lin2023xroute,hubrouter2023}. The former are limited by grid resolution, and the latter by the capability of the underlying router. PCBWorld~\citep{pcbworld2026} sought to overcome these limitations by letting an agent directly use the gridless routing functions of a real electronic design automation (EDA) engine, but its performance still drops sharply on complex boards. 

A key cause is the scarcity of training data for routing~\citep{lin2021async,gpcb2024}. First, routed boards are core intellectual property of individual companies and are rarely released. Public collections hold only about a thousand boards each~\citep{pcbench2024,omnirouting2026}. Second, in most cases, public boards contain only the completed board. The demonstration of how a designer operated the EDA tool is usually not recorded. Since many action sequences yield the same board, recovering the demonstration from a completed board is an ill-posed inverse problem. Finding an executable one is thus costly. Behavior cloning (BC) on action demonstrations followed by reinforcement learning (RL) fine-tuning has repeatedly succeeded across domains such as robot manipulation, games, and large language model (LLM) reasoning~\citep{alphago2016,dapg2018,instructgpt2022,deepseekr1_2025}. Unfortunately, no public dataset yet provides routing demonstrations that execute in an EDA engine.

To fill this gap, we propose \ours{} (Constructing Executable Demonstrations for Engine-grounded Training). To our knowledge, \ours{} is the first method in EDA routing to synthesize executable routing demonstrations at scale. It generates boards whose statistics resemble those of a reference board dataset, together with engine-level action demonstrations that route each board completely with zero design rule violations (DRVs). The key idea is inductive construction. \ours{} adds pads one at a time and commits a pad only when an engine-level action sequence connects it to the existing pads of its net with zero DRVs. Because each connection is added to an already routed board, the completed board is solvable by construction, and the recorded sequences form its demonstration. A best-first search tailored to routing finds these sequences efficiently. Finally, \ours{} re-executes every demonstration and keeps only those that again route the board completely with zero DRVs. As a result, \ours{} generates a large set of boards of comparable complexity from only a small reference board dataset. \ours{} then pretrains a routing policy on these demonstrations with BC and fine-tunes it with RL.

Our contributions are as follows.

\textbf{Method.}\quad We propose \ours{}, which synthesizes routing boards with executable demonstrations through inductive construction in an EDA engine and trains a routing policy on them with BC followed by RL fine-tuning.

\textbf{Data.}\quad We release \cadetdata{}, built with \ours{}, together with its generation code. It is the first large-scale dataset of routing demonstrations that execute in an EDA engine, with 100,000 synthetic routing instances. Its board statistics closely follow those of the reference board dataset.

\textbf{Results.}\quad On real test boards, \ours{} raised the clean pass (CP) rate, the fraction of boards routed completely with zero DRVs, from \rev{38.0\% to 67.2\%} over RL alone on \cadetdata{}. \rev{Its CP was comparable to that of the open-source autorouter FreeRouting~\citep{freerouting} in a third of the wall-clock time and higher with three rollouts in a similar time. It also exceeded the CP of all other baselines, including} GPT-6 Astra~\citep{openai2026gpt6astra}.

\section{Preliminaries}
\label{sec:preliminaries}

\subsection{PCB Routing}
\label{sec:pcb-routing}

A PCB electrically connects component contacts, called \emph{pads}, through copper traces. A \emph{net} specifies pads that must be electrically connected. Let $N$ be the number of nets, $P_i$ the number of pads in net $i$, and $P=\sum_{i=1}^{N}P_i$ the total pad count. Pad $p$ has location $\bm{x}_p$. Each net $i$ is assigned a \emph{netclass} from the board's netclass set $\mathcal C$. The netclass defines the net's rule vector $\bm{r}_i$, which includes track width $w_i$, clearance $c_i$, and via dimensions. Traces consist of track segments on copper layers, and \emph{vias} connect tracks across layers. A board state $s$ contains the board geometry, pads, net assignments, design rules, current routing, and the routing head, the position from which the next route segment starts. A routing instance is represented by its initial unrouted state $s_0$. Routing adds tracks and vias to connect every net while keeping distinct nets electrically isolated. A design rule check (DRC) reports DRVs of geometric constraints, such as minimum clearance. A successful solution completes all required connections with zero DRVs. Wirelength and via count characterize routing quality beyond this feasibility condition.

\subsection{Routing MDP and Demonstrations}
\label{sec:routing-mdp}

\begin{figure}[!htbp]
\centering
\includegraphics[width=\linewidth]{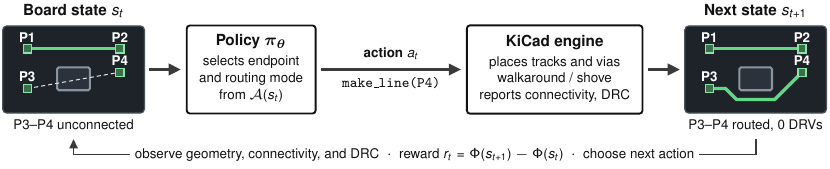}
\caption{\textbf{Routing with an EDA engine.}\quad At each step, the policy observes the board state $s_t$ and selects an action $a_t\in\mathcal A(s_t)$ that invokes a \kicad{} operation. The engine executes it and returns the updated board $s_{t+1}$ with connectivity and DRC results. In this example, the policy selects \texttt{make\_line} toward P4, and the engine routes the track around the obstacle. The reward $r_t$ is the resulting change in the potential $\Phi$.}
\label{fig:engine-interface}
\end{figure}

We use \pcbworld{}~\citep{pcbworld2026}, a Gymnasium environment~\citep{gymnasium2024} built on the \kicad{} EDA engine (\cref{fig:engine-interface}). At each step, the engine executes $a_t\sim\pi_\theta(\cdot\mid s_t)$ and produces $s_{t+1}=F(s_t,a_t)$, until completion or a step limit.

\textbf{State.}\quad
The state $s_t$ is a nested dictionary of board objects with explicit coordinates, dimensions, layers, and net assignments. It contains fixed pads and design rules, current tracks and vias, connectivity, and the routing head. The environment provides $s_t$ to LLM agents as S-expression text and to RL agents as indexed tokens.

\textbf{Action.}\quad
An engine-level action $a\in\mathcal A(s)$ specifies a native \kicad{} operation and its arguments, where $\mathcal A(s)$ is the candidate action set at state $s$. For example, \texttt{make\_line}($\bm{x}_a$) extends a route toward a requested endpoint $\bm{x}_a$. The \texttt{make\_via} operation additionally places a via to change layers. The policy selects a candidate endpoint and a routing mode, such as walkaround or push-and-shove. The engine constructs the resulting geometry (Appendix~\ref{app:environment} lists all actions).

\textbf{Reward.}\quad
The environment defines board quality through a configurable potential,
\begin{equation}
    \Phi(s)=\Phi_{\mathrm{con}}(s)-\bigl[f_{\mathrm{drv}}(s)+\lambda_w\ell(s)+\lambda_v n_{\mathrm{via}}(s)\bigr],
    \qquad r_t=\Phi(s_{t+1})-\Phi(s_t).
\end{equation}
Here, $f_{\mathrm{drv}}$ penalizes engine-reported DRVs, $\ell$ is total wirelength, and $n_{\mathrm{via}}$ counts vias, with nonnegative weights $\lambda_w,\lambda_v$. The term $\Phi_{\mathrm{con}}$ rewards connected and DRV-free nets and penalizes unrouted connections. Appendix~\ref{app:reward} gives all terms, their weights, and a small penalty in $r_t$ on invalid actions.

\textbf{Demonstration.}\quad
A demonstration is a trajectory $\tau=(s_0,a_0,s_1,\ldots,a_{T-1},s_T)$ of engine-level actions and resulting states. Many trajectories can reach the same $s_T$, so a completed board does not determine its demonstration. For an action sequence $\tau$, $F(s,\tau)$ denotes the state reached by executing $\tau$ in order from $s$. Replay reconstructs the states of $\tau$ by executing its actions from $s_0$. Fresh-engine replay does so on a newly started engine instance. Pads remain fixed throughout replay. Construction-time pad additions are outside this action space (\S\ref{sec:generation}).

\section{Method}
\label{sec:cadet}

\ours{} trains a routing policy in the environment of \S\ref{sec:routing-mdp}. Instead of recovering demonstrations from completed boards, it constructs synthetic PCB instances and engine-executable demonstrations simultaneously. The policy learns from replayed demonstrations through BC, followed by RL fine-tuning on the same instances.

\subsection{Board Attribute Distributions}
\label{sec:sampling-distributions}

The generator $\mathcal G_\psi$ constructs routable boards and demonstrations from sampled board attributes (Algorithm~\ref{alg:generation}, lines~\ref{line:sample-board}, \ref{line:sample-netclasses}, and~\ref{line:sample-nets}). These are pre-layout quantities that describe the board dimensions, netlist size, and design rules. Each attribute follows a simple parametric distribution or the empirical distribution of a reference dataset $\mathcal D_{\mathrm{ref}}$. We denote these distributions collectively by $q_\psi$, with parameters $\psi$ estimated from $\mathcal D_{\mathrm{ref}}$. Appendix~\ref{app:generation-details} details these distributions and additional pad attributes.

\textbf{Board dimensions.}\quad
Width $W$ and height $H$ define the board bounding box. The distribution $q_\psi(W,H)$ draws a board size $\sqrt{WH}$ and a long-to-short side ratio, which together determine $W$ and $H$. The size follows the lognormal distribution $\sqrt{WH}\sim\lognormal(\mu_{\mathrm{size}},\sigma_{\mathrm{size}}^2)$.

\textbf{Netlist size.}\quad
The net count $N$ is the nearest integer to $\rho_{\mathrm{net}}WH$, with the board area $WH$ in cm$^2$, where the net density $\rho_{\mathrm{net}}$ follows $\lognormal(\mu_{\mathrm{net}},\sigma_{\mathrm{net}}^2)$. This defines $q_\psi(N\mid W,H)$. The distribution $q_\psi(P_i)$ of each pad count $P_i\geq2$ is the empirical distribution of per-net pad counts.

\textbf{Design rules.}\quad
The netclass set $\mathcal C$ and each rule vector $\bm{r}_i$ (\S\ref{sec:pcb-routing}) are sampled empirically from $q_\psi(\mathcal C)$ and $q_\psi(\bm{r}_i\mid\mathcal C)$, so all rule values in $\bm{r}_i$ are drawn jointly.

\subsection{Board and Demonstration Generation}
\label{sec:generation}

\begin{figure}[!t]
\centering
\begin{subfigure}{\linewidth}
\centering
\includegraphics[width=\linewidth]{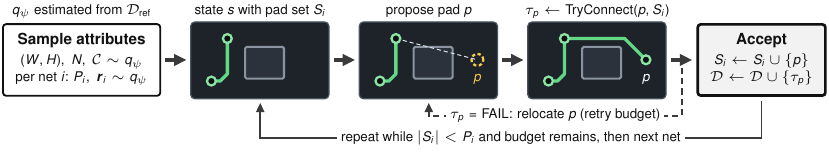}
\caption{Board and demonstration generation}
\label{fig:construction-gen}
\end{subfigure}\\[4pt]
\begin{subfigure}{\linewidth}
\centering
\includegraphics[scale=1]{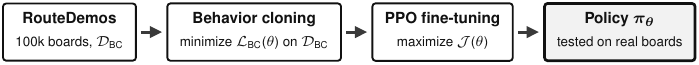}
\caption{Policy training}
\label{fig:construction-train}
\end{subfigure}
\caption{\textbf{Overview of \ours{}.}\quad (a) Inductive construction. The generator samples board attributes from $q_\psi$ and grows each net pad by pad. It accepts a proposed pad $p$ only when TryConnect connects $p$ to the pad set $S_i$ with zero DRVs, and it adds the action sequence $\tau_p$ to $\mathcal D$. Every accepted pad thus leaves all retained nets connected with zero DRVs. A failed proposal is relocated within a retry budget. (b) The policy is trained by BC on the replayed demonstrations and then fine-tuned with PPO.}
\label{fig:construction}
\end{figure}

\textbf{Inductive construction.}\quad
For each net $i$, the generator starts from a pad set $S_i=\{p_0\}$ with $p_0$ at a uniformly chosen location and adds pads one at a time, each connected to the net by a route with zero DRVs (Algorithm~\ref{alg:generation}, lines~\ref{line:init-pair}--\ref{line:commit-pad}, and \cref{fig:construction-gen}). Each subsequent pad is proposed near a randomly chosen existing pad of the net under a locality distribution (Appendix~\ref{app:generation-details}). The construction state $s$ holds all accepted pads, their routes, and the proposed pad. From $s$, TryConnect searches for an action sequence $\tau_p$ that connects the proposed pad to the net. Failed proposals are relocated within a retry budget. Once the budget is exhausted, the net ends with fewer than the intended $P_i$ pads. Because each pad is connected on an already routed board, the completed board is routable with high probability, and the sequences recorded in $\mathcal D$ form its candidate demonstration.

\begin{narrowalgorithm}[!t]{\linewidth}
\small
\caption{Constructing Routable Boards and Demonstrations}
\label{alg:generation}
\algtext*{EndFor}
\algtext*{EndWhile}
\algtext*{EndIf}
\begin{algorithmic}[1] 
\State Sample $(W,H)\sim q_\psi(\cdot,\cdot)$ and $N\sim q_\psi(\cdot\mid W,H)$ \label{line:sample-board}
\State Sample netclass set $\mathcal C\sim q_\psi(\cdot)$ \label{line:sample-netclasses}
\State Initialize demonstration set $\mathcal D\gets\varnothing$ and pad sets $S_i\gets\varnothing$ for all $i\in[N]$
\For{$i=1,\ldots,N$}
    \State Sample net $i$'s attributes: $P_i\sim q_\psi(\cdot)$ and $\bm{r}_i\sim q_\psi(\cdot\mid\mathcal C)$ \label{line:sample-nets}
    \State Propose initial pad $p_0$ \label{line:init-pair}
    \State $S_i\gets S_i\cup\{p_0\}$
    \While{$|S_i|<P_i$ and budget not exhausted}
        \State Propose pad $p$ \label{line:propose-pad}
        \State $\tau_p\gets\mathrm{TryConnect}(p,S_i)$ \label{line:try-connect}
        \If{$\tau_p\ne\mathrm{FAIL}$}
            \State $S_i\gets S_i\cup\{p\}, \, \mathcal D\gets\mathcal D\cup\{\tau_p\}$ \label{line:commit-pad}
        \EndIf
    \EndWhile
\EndFor
\State Form $s_0$ from the placed pads and set $\mathcal D_{\mathrm{BC}}\gets\mathrm{Replay}(s_0,\mathcal D)$ \Comment{Algorithm~\ref{alg:finalize}} \label{line:replay-bc}
\end{algorithmic}
\end{algorithm}
\begin{algorithm}[H]
\small
\caption{$\mathrm{TryConnect}(p,S_i)$: Best-First Connection Search}
\label{alg:connection-search}
\algtext*{EndWhile}
\algtext*{EndIf}
\begin{algorithmic}[1]
\Require Proposed pad $p$, pad set $S_i$ of net $i$, current state $s$
\State Initialize queue $Q\gets\{(\varnothing,a) : a\in \mathcal A(s)\}$
\While{$Q\ne\varnothing$ and budget not exhausted}
    \State Pop $(\tau_p,a)$ with the smallest $f(\tau_p,a)$ in Equation~\eqref{eq:search-cost} from $Q$ \label{line:select-candidate}
    \State $\tau_p\gets(\tau_p,a)$ and $s'\gets F(s,\tau_p)$ \label{line:execute-candidate}
    \If{$s'$ has zero DRVs}
        \If{$p$ is connected to $S_i$ in $s'$} \label{line:check-connection}
            \State \Return $\tau_p$
        \ElsIf{the routing head of $s'$ is unvisited}
            \State $Q\gets Q\cup\{(\tau_p,a') : a'\in \mathcal A(s')\}$ \label{line:expand-head}
        \EndIf
    \EndIf
\EndWhile
\State \Return FAIL
\end{algorithmic}
\end{narrowalgorithm}

\textbf{Connection search with TryConnect.}\quad
$\mathrm{TryConnect}(p,S_i)$ searches for an executable action sequence $\tau_p$ that routes $p$ to $S_i$ from the current state $s$ (Algorithm~\ref{alg:connection-search}). Each action $a\in\mathcal A(s)$ routes from the routing head of $s$ to its requested endpoint $\bm{x}_a$, which becomes the next routing head. Starting from $p$ with an empty $\tau_p$, the search appends actions one at a time. Each step selects the pair $(\tau_p,a)$ in the queue $Q$ with the smallest estimated cost $f(\tau_p,a)$ in Equation~\eqref{eq:search-cost}, where $a$ starts at the routing head that $\tau_p$ reaches. It executes this pair and returns $\tau_p$ once the resulting state has zero DRVs and connects $p$ to $S_i$ (lines~\ref{line:select-candidate}--\ref{line:check-connection}). Otherwise, a DRV-free state with an unvisited routing head adds $(\tau_p,a')$ to $Q$ for each candidate $a'$ from that head (line~\ref{line:expand-head}). It fails when the queue or expansion budget is exhausted.

The cost $f(\tau_p,a)$ combines the number of actions in $\tau_p$ with a grid-Dijkstra heuristic $h$ that estimates the distance from the requested endpoint of $a$ to $S_i$,
\begin{equation}
f(\tau_p,a)=2(|\tau_p|+1)+\bm{1}_{\mathrm{via}}(a)+
\begin{cases}
0, & \text{$\bm{x}_a$ lies on $S_i$},\\
h(\bm{x}_a), & \text{otherwise},
\end{cases}
\label{eq:search-cost}
\end{equation}
where $\bm{1}_{\mathrm{via}}(a)$ adds a unit penalty if $a$ is a via. Costs are estimated before execution in line~\ref{line:execute-candidate}.

After construction, we form $s_0$ from the placed pads of all nets and replay the action sequences in $\mathcal D$ in their recorded order to obtain the full trajectory $\tau$.  Replay keeps all retained pads fixed. Repair reroutes failed connections in $\tau$, and pruning removes pads from $s_0$ that repair cannot connect (Algorithm~\ref{alg:finalize}). The pair $(s_0,\tau)$ is accepted only when fresh-engine replay executes every action, connects all nets, and yields zero DRVs. Otherwise the board is discarded. The accepted pair yields the state--action pairs $\mathcal D_{\mathrm{BC}}$, written $\mathrm{Replay}(s_0,\mathcal D)$ in Algorithm~\ref{alg:generation} (line~\ref{line:replay-bc}).

\textbf{Pad-count calibration.}\quad
Exhausted retry budgets and pruning can make the realized total pad count $\hat P=\sum_i\hat P_i$ fall below the sampled total $P=\sum_i P_i$. To compensate, the net count is increased to $\bar N$, the nearest integer to $\alpha N$. The factor $\alpha$ is estimated in pilot runs by comparing the realized pad densities of generated boards with those of $\mathcal D_{\mathrm{ref}}$ (Appendix~\ref{app:generation-details}).

\subsection{Behavior Cloning and RL Fine-Tuning}
\label{sec:training}

BC trains the policy on the demonstrations of all accepted boards, each verified by fresh-engine replay (\cref{fig:construction-train}). It minimizes the negative log-likelihood of the recorded engine-level actions over their state--action pairs $\mathcal D_{\mathrm{BC}}$,
\begin{equation}
\mathcal{L}_{\mathrm{BC}}(\theta)
= -\mathbb{E}_{(s,a)\sim\mathcal D_{\mathrm{BC}}}\log \pi_{\theta}(a\mid s).
\end{equation}
BC learns from recorded decisions without optimal-action labels. Multiple routing sequences can solve the same board, and a successful demonstration need not maximize the routing reward.

The resulting policy $\pi_{\theta_{\mathrm{BC}}}$ initializes PPO fine-tuning on the same boards. Fine-tuning maximizes the PPO clipped surrogate objective $\mathcal J_{\mathrm{PPO}}$~\citep{ppo2017} with a KL penalty toward the BC policy,
\begin{equation}
\mathcal J(\theta)=\mathcal J_{\mathrm{PPO}}(\theta)-\beta\,\mathbb E_{s}\bigl[D_{\mathrm{KL}}\bigl(\pi_\theta(\cdot\mid s)\,\|\,\pi_{\theta_{\mathrm{BC}}}(\cdot\mid s)\bigr)\bigr],
\label{eq:ppo-kl}
\end{equation}%
with $\beta=0.05$ in the main configuration. Subsequent interaction allows the policy to improve under the routing objective.

We call the policy before fine-tuning \ours{} (BC). Scratch PPO uses random initialization on the same board dataset. BC does not train the critic, so fine-tuning begins with a critic warm-up. It also detaches the critic from the shared encoder. The comparison therefore evaluates the full BC and fine-tuning procedure, including its optimization settings. Generation and BC also incur costs beyond the RL interaction budget.

\section{Experiments}
\label{sec:experiments}

\begin{figure}[t]
\centering
\newlength{\figw}\setlength{\figw}{\textwidth}%
\newcommand{\boardslot}[2]{\parbox[c][0.19\figw][c]{#1\figw}{\centering\includegraphics[width=#1\figw,height=0.19\figw,keepaspectratio]{figures/boards/#2}}\\[6pt]}%
\begin{subfigure}[t]{0.33\textwidth}
\centering
\boardslot{0.33}{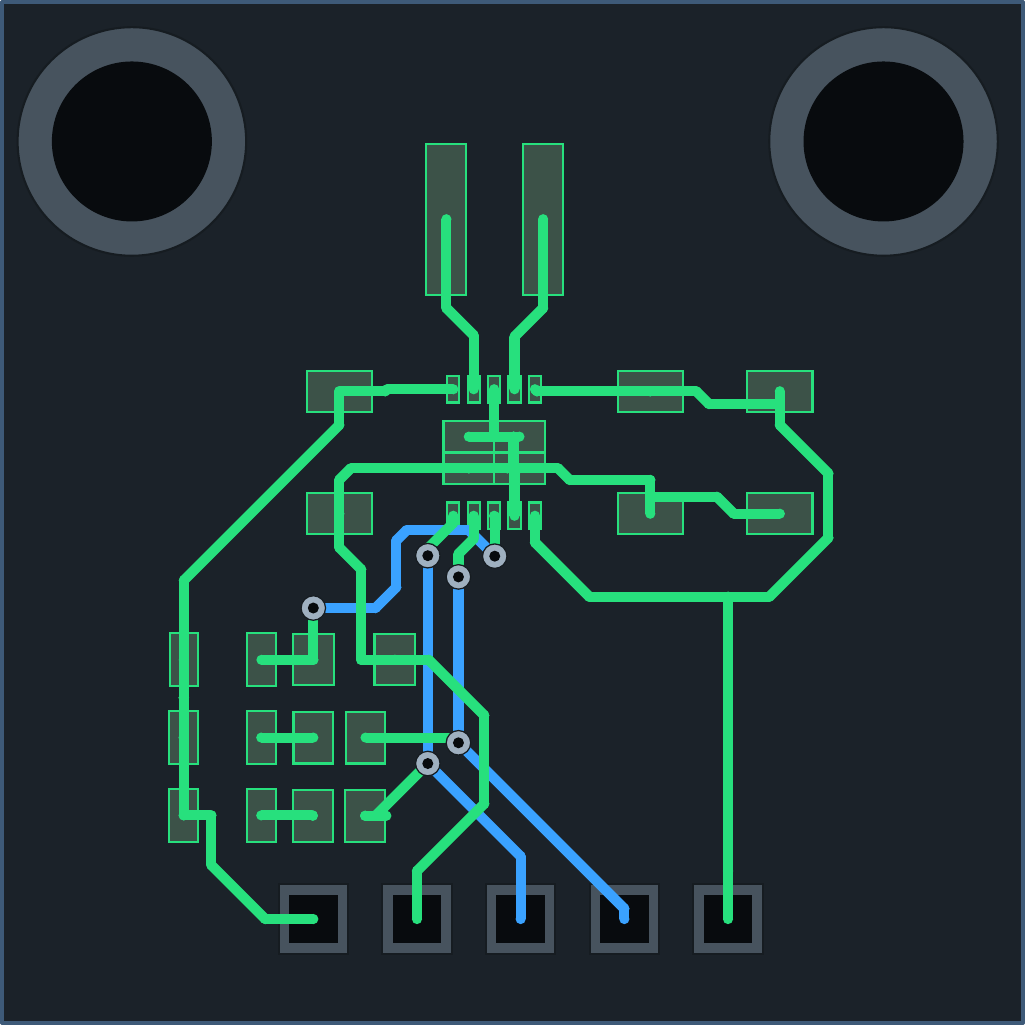}
\boardslot{0.33}{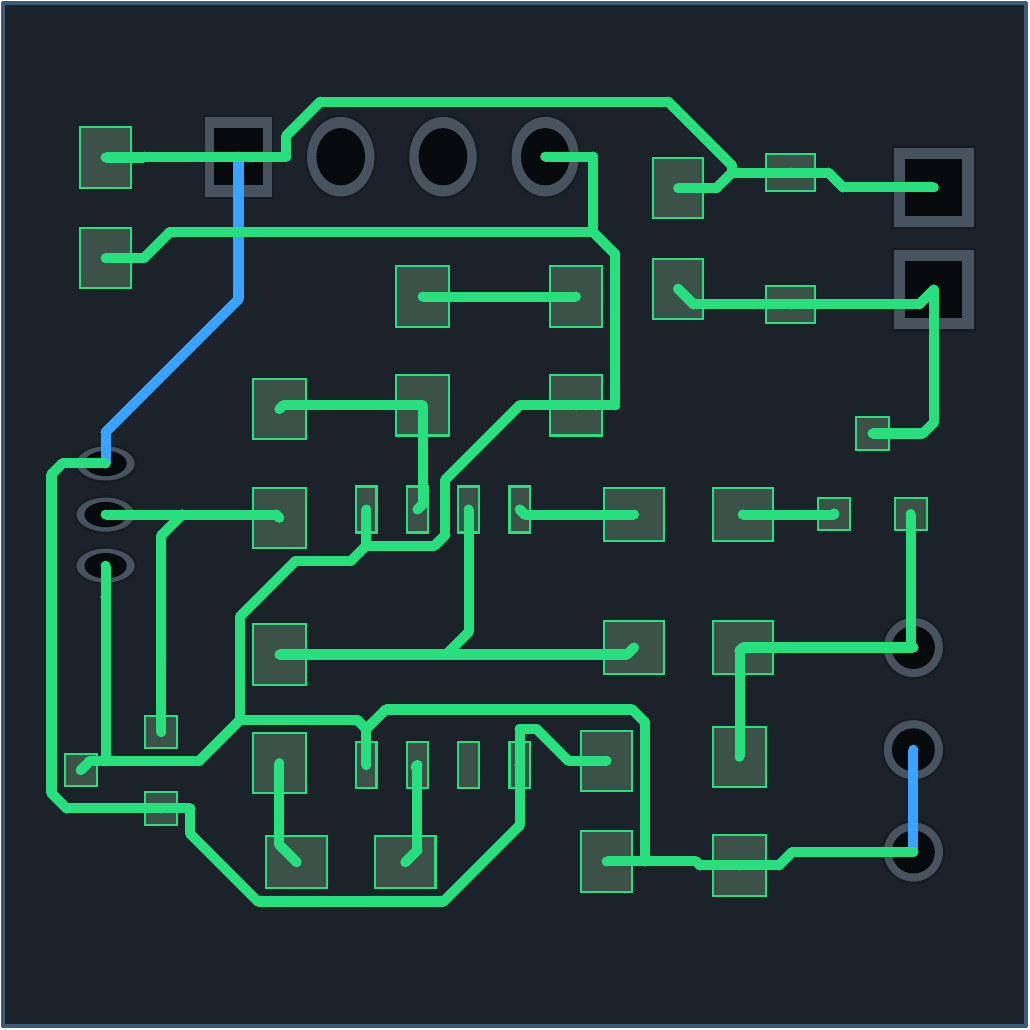}
\boardslot{0.33}{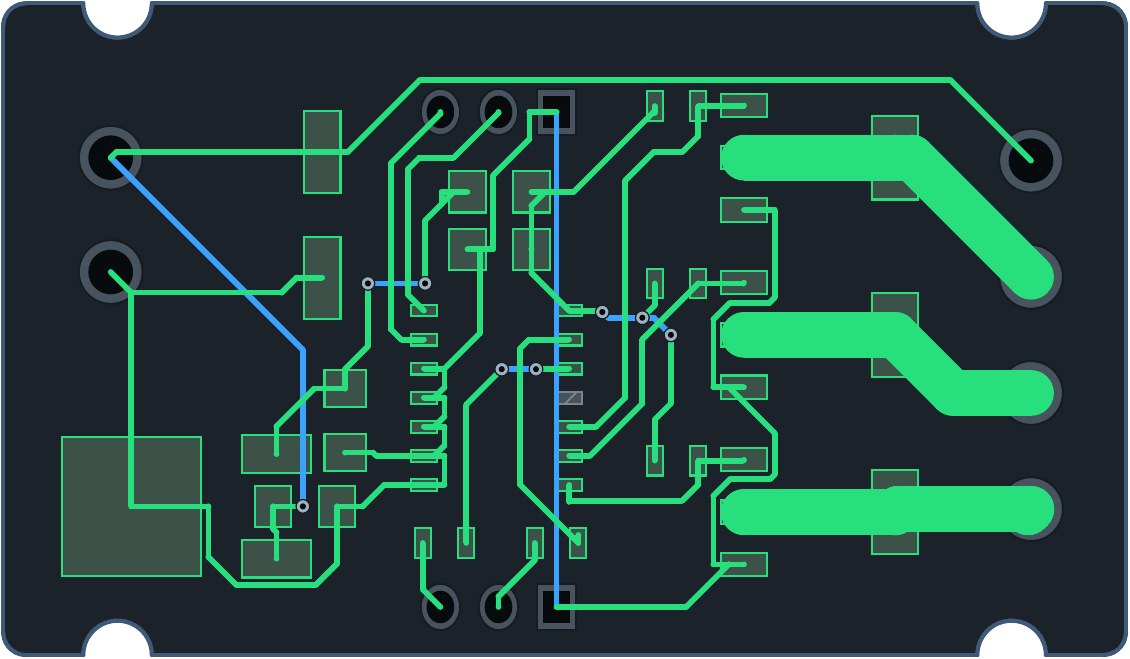}
\caption{D3-B test}
\end{subfigure}\hspace{0.05\textwidth}%
\begin{subfigure}[t]{0.33\textwidth}
\centering
\boardslot{0.33}{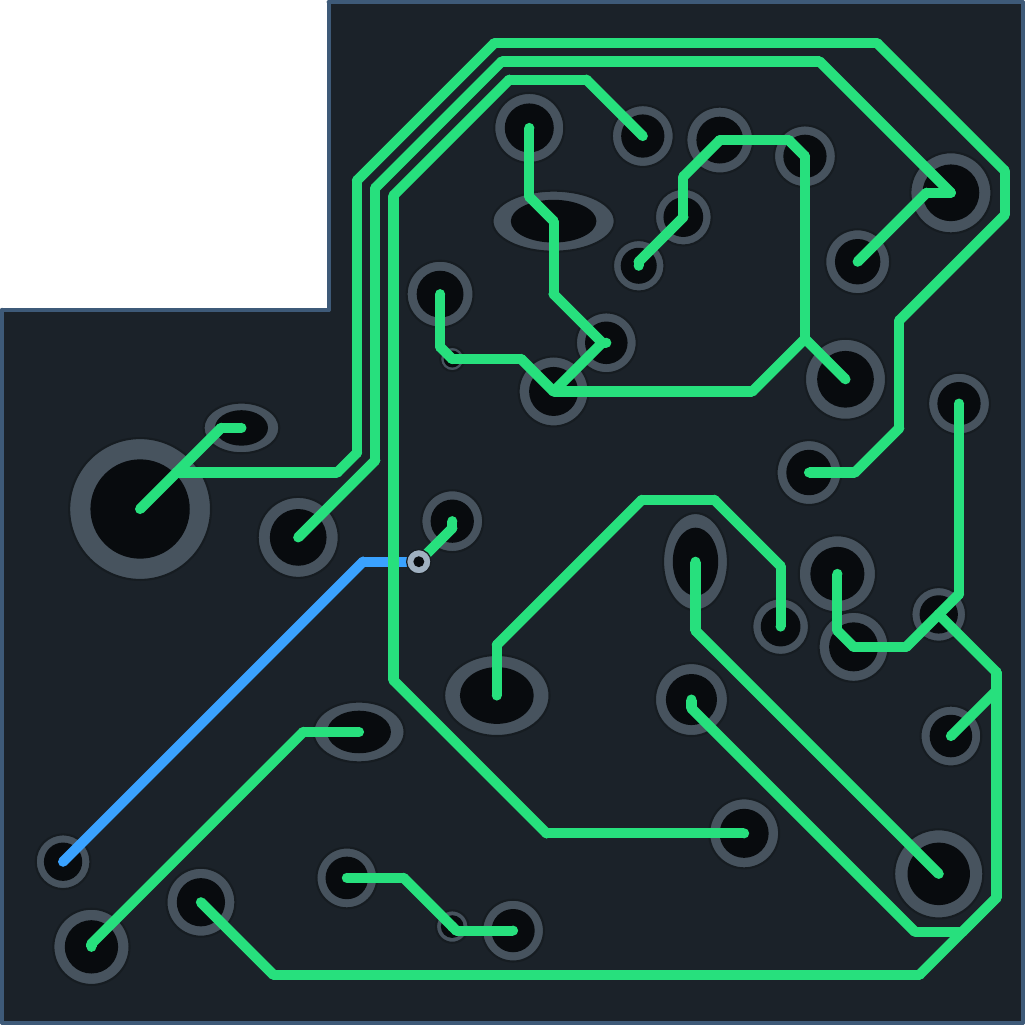}
\boardslot{0.33}{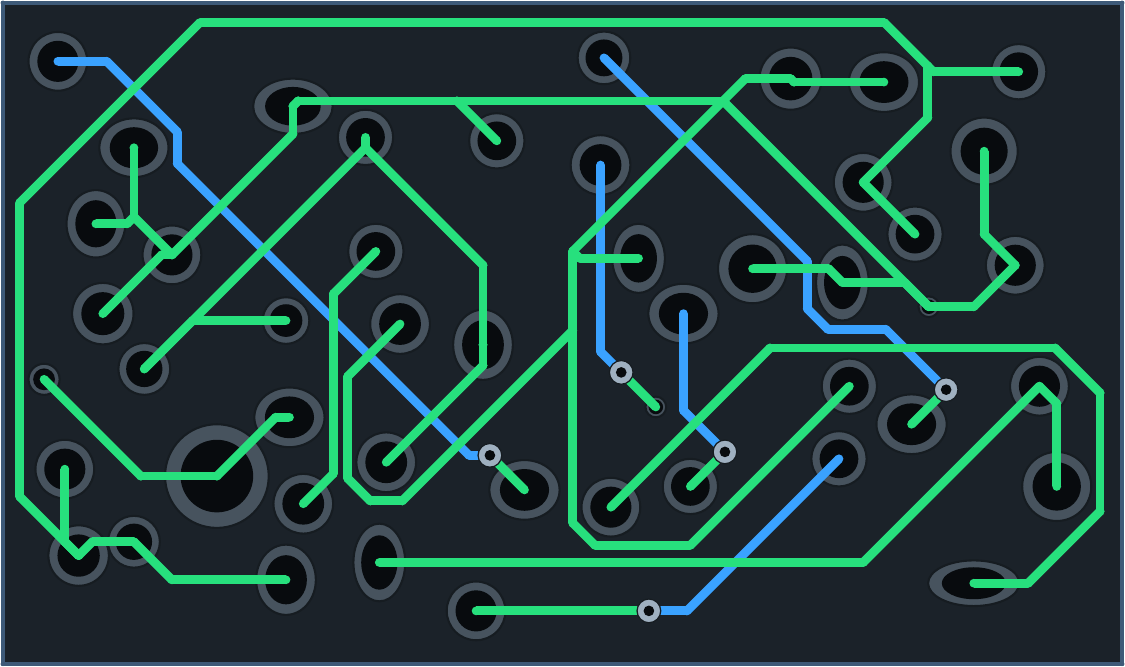}
\boardslot{0.33}{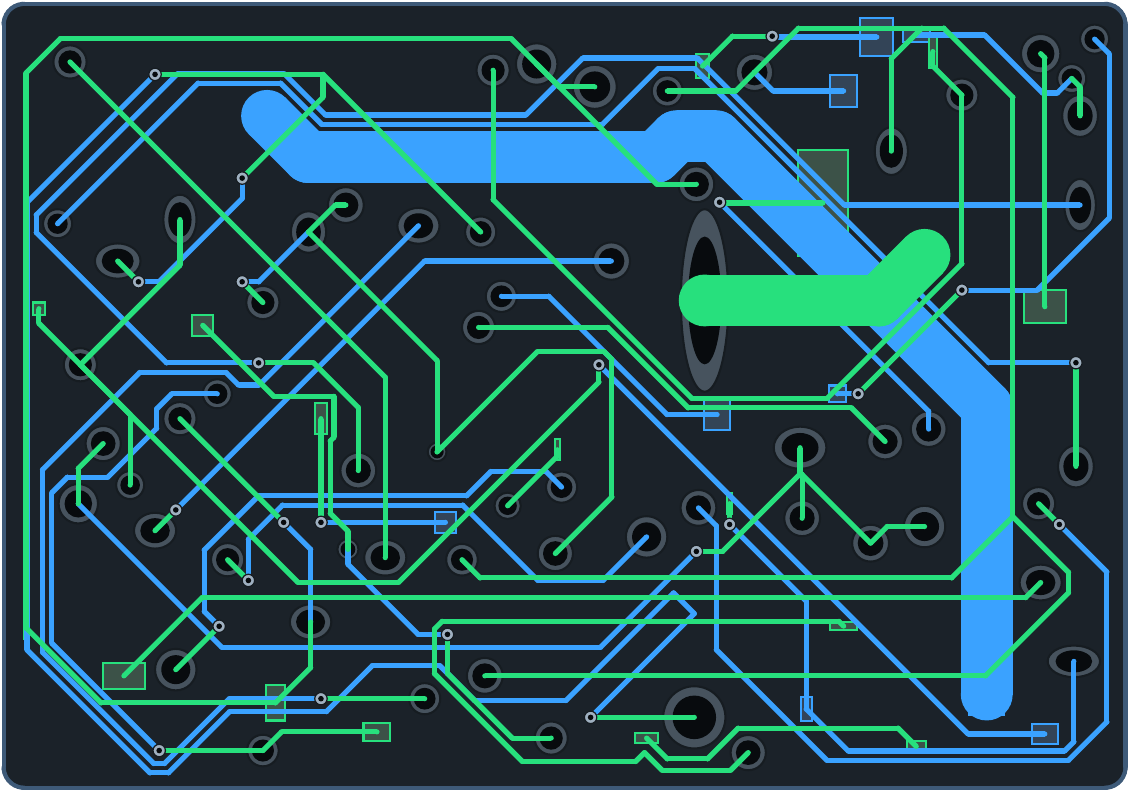}
\caption{\cadetdata{}}
\end{subfigure}\hspace{0.05\textwidth}%
\begin{subfigure}[t]{0.21\textwidth}
\centering
\boardslot{0.21}{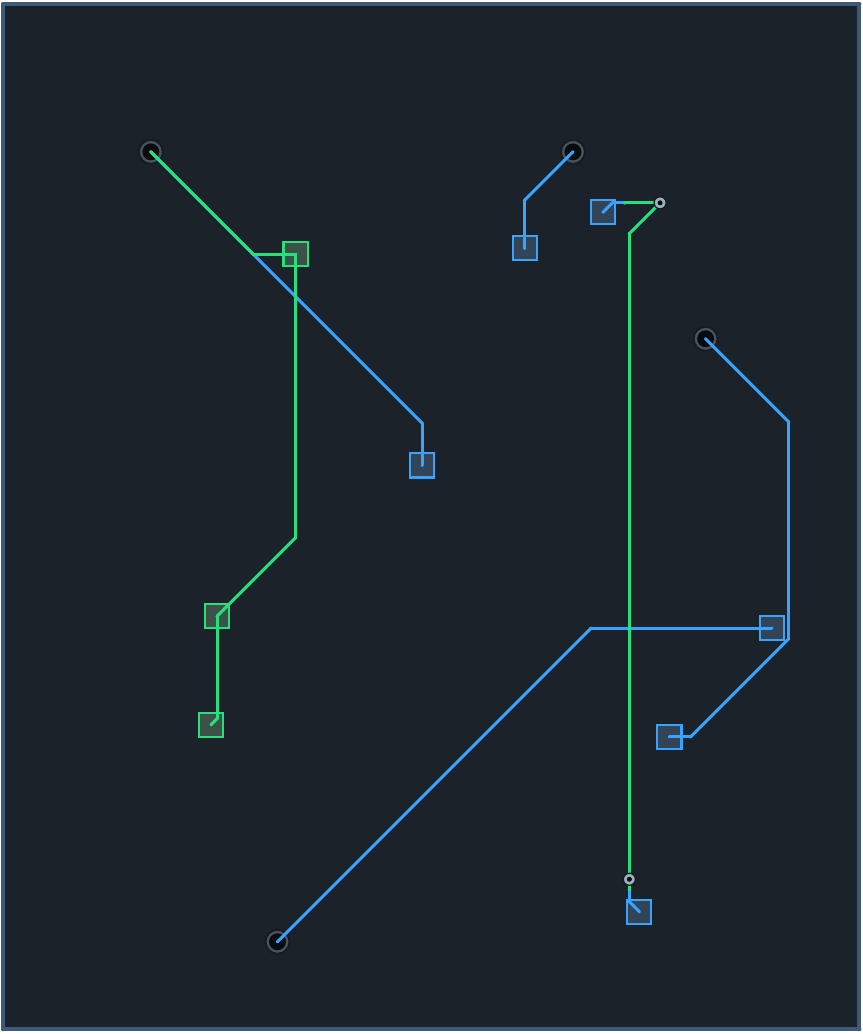}
\boardslot{0.21}{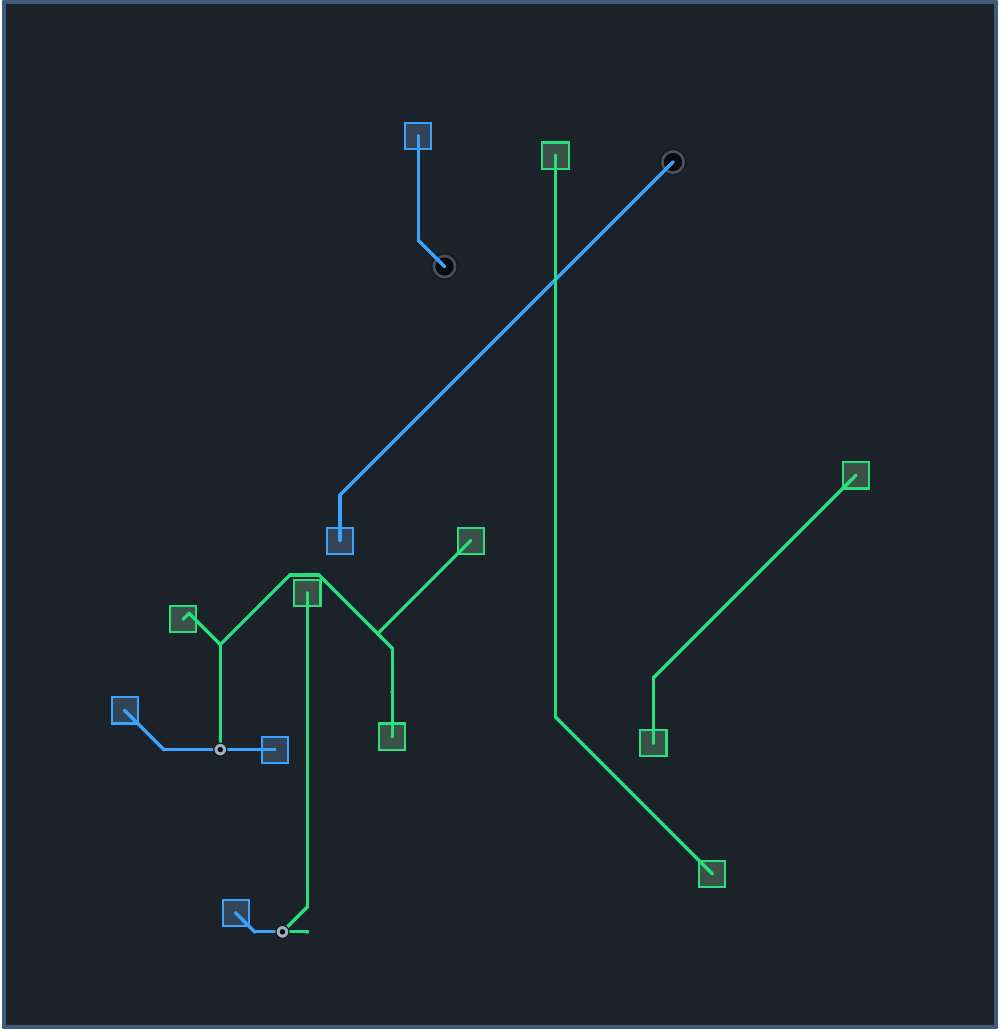}
\boardslot{0.21}{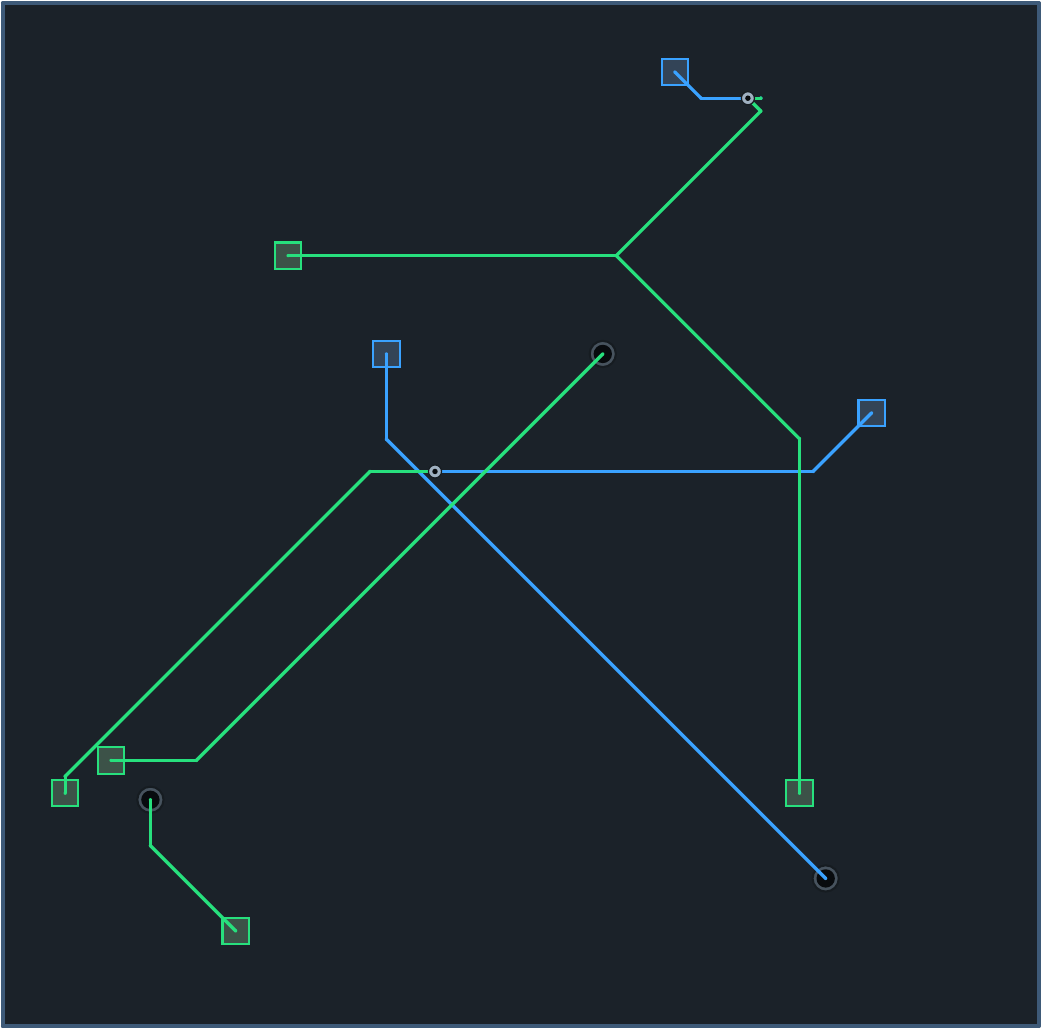}
\caption{D2}
\end{subfigure}
\caption{Example routed boards from D3-B test, \cadetdata{}, and D2. Green and blue traces lie on different copper layers.}
\label{fig:board-examples}
\end{figure}

\subsection{Experimental Setup}
\label{sec:setup}

\textbf{Data.}\quad
We evaluated routing policies on two real-board sets from the PCBWorld benchmark~\citep{pcbworld2026}, D3-A with 6 to 31 pads per board and D3-B with 31 to 100.
We used 99 D3-A boards for evaluation only.
From the 286 D3-B boards, we kept the 283 two-layer boards and split them into D3-B ref (142 boards) and D3-B test (141).
\ours{} used D3-B ref as the reference dataset $\mathcal D_{\mathrm{ref}}$ to produce \cadetdata{}, 100,000 boards with executable demonstrations, about 700 times the size of $\mathcal D_{\mathrm{ref}}$.
The PCBWorld synthetic training set D2 provides 10,000 boards with 8 to 21 pads each.

\textbf{Baselines.}\quad
The rule-based baselines were the open-source routers FreeRouting~\citep{freerouting} and KiCad Routing Tools~\citep[KRT;][]{kicadroutingtools}.
The LLM baselines, GPT-6 Astra~\citep{openai2026gpt6astra} and GPT-6 Luna~\citep{openai2026gpt6solluna}, routed boards by interacting with the PCBWorld environment (settings and prompts in Appendix~\ref{app:llm-prompt}).
PPO ($X$) denotes a policy trained from random initialization on dataset $X$.
PPO (D2) reimplements the PCBWorld policy with our training configuration and exceeds the original in CP on D3-A.
\cref{tab:main-results} labels it PCBWorld (PPO).

\textbf{Evaluation.}\quad
Every method was scored on its final \kicad{} board file with the same post-hoc DRC.
CP is the fraction of boards with all connections completed and zero error-severity DRVs.
Routability (Rout.) is the fraction of required connections completed.
The potential gain $\Delta\Phi=\Phi(s_T)-\Phi(s_0)$ measures improvement over the unrouted board, with $\Phi$ from \S\ref{sec:routing-mdp}.
FreeRouting \rev{ran four times per board.}\footnote{KRT is deterministic and LLM inference is costly, so each ran once.}
\rev{Each learned policy was trained with four seeds (42 to 45), and each seed's policy ran four times per board.}
Learned policies used a 1,024-step horizon.
For any method, @$k$ keeps the best of $k$ rollouts in each run, by CP and then $\Delta\Phi$.
Tables report means over these runs\rev{, and over the four seeds for learned policies}. Appendix~\ref{app:metrics} defines all metrics formally, and Appendix~\ref{app:full-results} adds standard deviations, \rev{95\% confidence intervals,} wirelength, via counts, and DRVs.
Time is the mean wall-clock time per board on a single server with an AMD EPYC 9354 CPU and one NVIDIA L40 GPU.

\textbf{Training.}\quad
All policies used the same six-layer Transformer and PPO hyperparameters.
\ours{} applied BC and then 200 PPO fine-tuning iterations with $\beta=0.05$, critic warm-up, and critic detachment.
Scratch PPO ran 500 iterations.
Appendix~\ref{app:training-details} lists the architecture, hyperparameters, and configuration differences.
For each run, we selected the checkpoint with the highest mean $\Delta\Phi$ on D3-B ref.
Appendix~\ref{app:reproducibility} provides the source code, software versions, compute, and baseline settings.

\subsection{Analysis of \cadetdata{}}
\label{sec:data-properties}

\textbf{Acceptance rate and runtime.}\quad
We analyzed the Algorithm~\ref{alg:generation} runs that produced \cadetdata{}.
A pad that TryConnect fails to connect is relocated and, once the retry budget is exhausted, dropped from its net without discarding the board.
Replay with repair and pruning (line~\ref{line:replay-bc}) accepted 99.1\% of the boards after fresh-engine replay.
This high rate reflects that inductive construction keeps each board solvable before replay.
The full pipeline takes about 146 hours on one 56-core CPU server.
Appendix~\ref{app:generation-cost} gives full details.

\begin{table}[!t]
\centering
\small
\caption{Marginal KS distances to D3-B ref over 17 statistics, used to calibrate the generator. Lower is closer. Bold marks the lower distance between the two synthetic datasets. D3-B test provides a reference between real splits, not a lower bound. SMD and THT denote surface-mount and through-hole pads. Appendix~\ref{app:statistics} defines each statistic.}
\label{tab:fidelity}
\setlength{\tabcolsep}{0pt}%
\begin{tabular*}{\textwidth}{@{\extracolsep{\fill}}lrrr@{\hspace{1.5em}}lrrr}
\toprule
Statistic & D3-B test & \cadetdata{} & D2 & Statistic & D3-B test & \cadetdata{} & D2 \\
\midrule
Board size & 0.064 & \textbf{0.075} & 0.971 & THT drill & 0.108 & \textbf{0.058} & 0.970 \\
Aspect ratio & 0.170 & \textbf{0.072} & 0.504 & Clearance & 0.045 & \textbf{0.018} & 0.933 \\
Net density & 0.071 & \textbf{0.107} & 1.000 & Track width & 0.059 & \textbf{0.045} & 0.630 \\
Pad density & 0.057 & \textbf{0.139} & 1.000 & Netclasses & 0.028 & \textbf{0.129} & 0.660 \\
Locality & 0.026 & \textbf{0.066} & 0.392 & THT fraction & 0.057 & \textbf{0.026} & 0.444 \\
Pads per net & 0.015 & \textbf{0.016} & 0.108 & Front SMD fraction & 0.044 & \textbf{0.037} & 0.728 \\
SMD short side & 0.078 & \textbf{0.024} & 0.985 & Nets per board & 0.050 & \textbf{0.236} & 0.986 \\
SMD long side & 0.087 & \textbf{0.019} & 0.907 & Pads per board & 0.021 & \textbf{0.245} & 1.000 \\
THT pad size & 0.106 & \textbf{0.149} & 0.917 & Mean of all & 0.064 & \textbf{0.086} & 0.773 \\
\bottomrule
\end{tabular*}
\end{table}

\textbf{Quantitative similarity.}\quad
\cref{tab:fidelity} reports two-sample Kolmogorov--Smirnov (KS) distances~\citep{smirnov1948} to D3-B ref, used for generator calibration. \cadetdata{} has lower KS distances than D2 on all 17 statistics, and its mean KS distance of 0.086 is 89\% lower than the 0.773 for D2 and close to the 0.064 for D3-B test. Appendix~\ref{app:statistics} reports all 17 distances to D3-B test and the sample sizes.

\textbf{Qualitative similarity.}\quad
\cref{fig:board-examples} shows example boards from each dataset.
\cadetdata{} not only provides engine-level demonstrations but also contains boards much closer to real boards than D2 in shape and complexity.
D2 boards are smaller and sparser.
\cadetdata{} samples design rules empirically from reference boards and thus reproduces their diverse track widths.

\subsection{Main Results}
\label{sec:main-results}

\begin{table}[!t]
\centering
\caption{Evaluation on real test boards. Reference is the original human layout. Bold marks the best value among the other methods. \rev{Learned policies are averaged over four training seeds.} CP and Rout. are in percent, and Time is in seconds per board.}
\label{tab:main-results}
\small
\setlength{\tabcolsep}{2.8pt}
\begin{tabular}{ll!{\hskip 4pt}rrrr!{\hskip 5pt}rrrr}
\toprule
& & \multicolumn{4}{c!{\hskip 5pt}}{D3-B test (141)} & \multicolumn{4}{c}{D3-A (99)} \\
\cmidrule(l{0pt}r{5pt}){3-6}\cmidrule(l{0pt}r{0pt}){7-10}
Method & Type & CP & $\Delta\Phi$ & Rout. & Time & CP & $\Delta\Phi$ & Rout. & Time \\
\midrule
Reference & -- & 100.0 & 144.41 & 100.0 & -- & 100.0 & 50.61 & 100.0 & -- \\
\midrule
FreeRouting @1 & Rule-based & 65.6 & 119.33 & \textbf{96.8} & 45.6 & 82.1 & 46.05 & 97.7 & 8.6 \\
KRT @1 & Rule-based & 27.7 & 84.71 & 78.3 & \textbf{2.7} & 69.7 & 40.54 & 92.8 & \textbf{1.5} \\
GPT-6 Astra @1 & LLM & 56.0 & 104.20 & 86.0 & 491.7 & 83.8 & 45.03 & 93.8 & 138.9 \\
GPT-6 Luna @1 & LLM & 5.0 & 37.72 & 47.9 & 335.7 & 46.5 & 31.30 & 78.5 & 179.1 \\
PCBWorld (PPO) @1 & RL & \rev{40.4} & \rev{80.66} & \rev{74.4} & -- & \rev{78.4} & \rev{42.28} & \rev{90.6} & -- \\
\midrule
\ours{} @1 & BC+RL & \rev{67.2} & \rev{107.22} & \rev{91.7} & \rev{15.5} & \rev{89.5} & \rev{45.13} & \rev{96.6} & \rev{4.2} \\
\ours{} @3 & BC+RL & \rev{\textbf{79.8}} & \rev{\textbf{121.30}} & \rev{95.9} & \rev{47.6} & \rev{\textbf{96.3}} & \rev{\textbf{48.87}} & \rev{\textbf{99.2}} & \rev{12.3} \\
\bottomrule
\end{tabular}
\end{table}

\rev{\Cref{tab:main-results} shows that \ours{} @1 exceeded every baseline in CP on D3-A. On D3-B test, its CP of 67.2\% was comparable to the 65.6\% of FreeRouting and above every other baseline.}
This result shows the effectiveness of BC on engine-level demonstrations that guarantee a clean pass\rev{, followed by RL fine-tuning}.
\ours{} @1 also exceeded every other agent in the same environment on every metric \rev{of \cref{tab:main-results}}.
The 6.79M-parameter \ours{} @1 outperformed the frontier GPT-6 Astra and ran about \rev{32} times faster.
Its margin over the lower-priced GPT-6 Luna was far wider.
PCBWorld (PPO) uses the same architecture without BC. \ours{} @1 exceeded it by \rev{26.8 percentage points} in CP and by \rev{26.56} in $\Delta\Phi$ on D3-B test, whose boards have more pads, and by \rev{11.1 points} in CP on D3-A.

Among the rule-based routers, KRT ran fastest but reached only 27.7\% CP on D3-B test.
\rev{FreeRouting was the strongest baseline. On D3-B test, \ours{} @1 had lower routability and $\Delta\Phi$ than FreeRouting at a comparable CP, but ran in a third of the wall-clock time, 15.5 seconds against 45.6.}
\rev{In a comparable 47.6 seconds, \ours{} @3 outperformed it by 14.2 points in CP. Its routability was comparable, 95.9\% against 96.8\%, but \cref{tab:full-results} shows that it used 9.1 vias per board against 4.1.}
On D3-A, \ours{} @3 nearly matched the reference, with \rev{96.3}\% CP and \rev{99.2}\% routability.

\subsection{Where Does the Gain Come From?}
\label{sec:ablation}

\begin{table}[htbp]
\centering
\caption{Effect of demonstrations and training data. \ours{} is \ours{} @1 of \cref{tab:main-results}. Bold marks the best value in each column. CP and Rout. are in percent.}
\label{tab:training-data}
\small
\setlength{\tabcolsep}{5pt}
\begin{tabular}{l!{\hskip 6pt}rrr!{\hskip 8pt}rrr}
\toprule
& \multicolumn{3}{c!{\hskip 8pt}}{D3-B test (141)} & \multicolumn{3}{c}{D3-A (99)} \\
\cmidrule(l{0pt}r{8pt}){2-4}\cmidrule(l{0pt}r{0pt}){5-7}
Method & CP & $\Delta\Phi$ & Rout. & CP & $\Delta\Phi$ & Rout. \\
\midrule
\ours{} & \rev{\textbf{67.2}} & \rev{\textbf{107.22}} & \rev{91.7} & \rev{\textbf{89.5}} & \rev{\textbf{45.13}} & \rev{\textbf{96.6}} \\
\ours{} (BC) & \rev{43.1} & \rev{78.29} & \rev{81.7} & \rev{83.2} & \rev{41.61} & \rev{94.4} \\
PPO (\cadetdata{}) & \rev{38.0} & \rev{95.98} & \rev{\textbf{92.0}} & \rev{80.2} & \rev{44.74} & \rev{\textbf{96.6}} \\
PPO (\cadetdata{} w/o TryConnect) & \rev{30.9} & \rev{90.54} & \rev{91.1} & \rev{76.6} & \rev{43.27} & \rev{96.3} \\
PPO (D2) & \rev{40.4} & \rev{80.66} & \rev{74.4} & \rev{78.4} & \rev{42.28} & \rev{90.6} \\
PPO (D3-B ref) & \rev{38.3} & \rev{94.87} & \rev{91.4} & \rev{75.6} & \rev{42.46} & \rev{95.7} \\
\bottomrule
\end{tabular}
\end{table}

\cref{tab:training-data} traces the gain of \ours{} over PPO (D2) in two steps, changing the training boards and then adding demonstrations.

\textbf{Effect of training boards.}\quad
\rev{On D3-B test, PPO (D2), PPO (D3-B ref), and PPO (\cadetdata{}) had similar CP, between 38\% and 40\%.
In $\Delta\Phi$ and routability, PPO (\cadetdata{}) and PPO (D3-B ref) were close and far above PPO (D2), which PPO (\cadetdata{}) exceeded by 15.32 and 17.6 points.
Training on \cadetdata{}, which uses only the attribute distributions of D3-B ref, thus matched training on the real boards.}
The variant \cadetdata{} w/o TryConnect only places pads, without connecting them by TryConnect, so its boards are not guaranteed to admit a clean pass.
PPO (\cadetdata{} w/o TryConnect) \rev{had lower means than} PPO (\cadetdata{}) on every metric\rev{, though within the variation across seeds}.
Because TryConnect also determines which pads are accepted, this comparison assesses the construction procedure jointly.

\textbf{Effect of demonstrations.}\quad
\rev{Most of the gain came from the procedure that \ours{} proposes, BC on demonstrations followed by RL fine-tuning.}
Although both used the same training boards, \ours{} exceeded PPO (\cadetdata{}) on D3-B test by \rev{29.2 points} in CP and by \rev{11.24} in $\Delta\Phi$.
\rev{Trained by BC alone on demonstrations that all end in a clean pass, \ours{} (BC) already reached 43.1\% CP, above the 38.0\% of PPO (\cadetdata{}) though within the variation across seeds, and RL fine-tuning added a further 24.1 points.}
This result shows that, as in other sequential decision problems~\citep{dapg2018}, executable demonstrations, which public boards lack, make BC \rev{followed by RL fine-tuning} effective for engine-level routing, beyond what synthetic boards alone provide.
\rev{Fine-tuning also differs from scratch PPO in the KL penalty and critic detachment (\S\ref{sec:training}).
In \cref{tab:finetune-ablation} of Appendix~\ref{app:full-results}, with either setting or both turned off, CP stayed between 63.3\% and 66.5\%, against 38.0\% for PPO (\cadetdata{}). The two settings are thus not the main source of the gain.}

\rev{The demonstrations also raised the via count. In \cref{tab:full-results,tab:finetune-ablation}, \ours{} (BC) used 11.9 vias per board and \ours{} 10.4, or 7.7 when fine-tuned without the KL penalty, against 4.8 for PPO (\cadetdata{}).}
\rev{\ours{} (BC) had lower D3-B test routability than PPO (\cadetdata{}), and fine-tuning closed most of this gap.}
We attribute this to demonstrations that all end in a clean pass. The policy keeps working on a difficult net instead of abandoning it, which raises CP but can leave other nets unrouted.
Fine-tuning improved every metric over \ours{} (BC) on both evaluation sets.
Offline RL on demonstrations that include some unclean ones is a promising direction for reducing this trade-off.

\section{Related Work}
\label{sec:related-work}

\textbf{Rule-based routers.}\quad
PCB autorouters such as FreeRouting~\citep{freerouting} and KRT~\citep{kicadroutingtools} combine path search with rip-up and rerouting.
They remain the practical standard but have difficulty completing complex, dense boards.
Integrated circuit (IC) routers~\citep{cugr,tritonroute} operate under different layer counts, design rules, and routing primitives.

\textbf{Learning-based routing.}\quad
Learned routers act at the grid, router, or engine level.
Grid-level policies move a route one cell at a time, learning from reward~\citep{liao2020globalrouting,circuitrouting2022mcts,jumanji} or from maze-router and manual routes~\citep{hebao2020}.
Their action set is small, but real boards need a fine grid resolution, which makes the search impractically large~\citep{pcbworld2026}.
Router-level policies delegate path construction to a rule-based router and choose only some of its inputs.
These inputs include net orderings~\citep{lin2021async,lin2023xroute,netordering2025transformer,liao2026dreamer}, router cost weights~\citep{khan2025offline}, fanout endpoints~\citep{fanoutnet2023}, and coarse routing guidance~\citep{geniusroute2019,hubrouter2023,neuralsteiner2024,gpcb2024,dsbrouter2025}.
Some learn from reward and others from routes produced by people or solvers.
The policy cannot correct paths that the router fails to build, so performance depends heavily on the underlying router.
Engine-level policies act through the routing operations of an EDA engine.
PCBWorld~\citep{pcbworld2026} exposes native \kicad{} routing operations as actions.
This intermediate level lets the policy shape route geometry with far fewer steps than grid moves.
However, current engine-level policies remain limited on complex boards.
To our knowledge, no prior work trains an engine-level routing policy from demonstrations.

\textbf{Routing datasets.}\quad
PCBench~\citep{pcbench2024} and OmniRouting~\citep{omnirouting2026} collect real designs, and PCBWorld~\citep{pcbworld2026} builds its real-board benchmarks from PCBench.
These resources hold at most completed boards, not demonstrations.
Synthetic generators from PCBWorld and Jumanji provide unrouted instances~\citep{jumanji,pcbworld2026}.
Grid generators lack a feasibility guarantee~\citep{liao2020globalrouting}, store only final Steiner trees~\citep{respack2022}, or discard the solving walks~\citep{jumanji}.
To our knowledge, no public real or synthetic dataset stores routing actions that replay in an EDA engine.

\section{\rev{Limitations}}
\label{sec:limitations}

\rev{Our evaluation covered two-layer boards with at most 100 pads. Boards with hundreds of pads or with more than two layers were not evaluated. Only three of the 286 D3-B boards have more than two layers, too few for the generator to fit a multilayer distribution. On D3-B test, \ours{} @1 was faster than FreeRouting at a comparable CP, but it still had lower routability and used more vias (\cref{tab:full-results}).}

\section{Conclusion}
\label{sec:conclusion}

We proposed \ours{} to address a key bottleneck of learning-based routing, the scarcity of demonstrations that execute in an EDA engine. \ours{} generates engine-level demonstrations and trains a routing policy on them with BC followed by RL fine-tuning. Its inductive construction adds pads one at a time, each with an engine-level action sequence that connects it to its net. By maintaining zero DRVs at every step, it efficiently produces complex DRV-free boards with the desired statistics. \ours{} estimates the distributions of key board attributes from real reference boards, and the resulting \cadetdata{} closely matches these boards in 17 statistics. On real test boards, the trained policy \rev{raised CP from 38.0\% to 67.2\% over RL alone on the same boards. On D3-B test, its CP of 67.2\% was comparable to the 65.6\% of FreeRouting in a third of the wall-clock time, and it reached 79.8\% with three rollouts in a similar time. It also exceeded the other baselines in CP.} Our analysis shows the potential of DRV-free executable demonstrations for learned EDA layout. It also motivates more refined demonstration generation and the collection of expert demonstrations from human engineers.

\subsection*{AI Use Statement}

We used LLMs (Claude Opus 5.5 and GPT-6 Astra) to write code, draft and polish sections of the paper, search for related work, and discuss research ideas. The authors reviewed all generated code, checked every cited work, and revised all text. The GPT-6 Astra and GPT-6 Luna agents in \S\ref{sec:setup} are baselines evaluated through their APIs. We take responsibility for the final content of this work, including text, claims, and artifacts.

\bibliography{references}
\bibliographystyle{iclr2027_conference}

\clearpage
\appendix
\section{Generated-Board Statistics}
\label{app:statistics}

\cref{tab:ks-full} reports all 17 marginal KS distances to D3-B test boards. It complements the D3-B ref comparison in \cref{tab:fidelity}. The Kolmogorov--Smirnov (KS) distance is the maximum absolute difference between empirical cumulative distribution functions. Smaller values indicate closer distributions.

The statistical analysis included 142 D3-B ref boards, 141 D3-B test boards, and 10,000 D2 boards. For \cadetdata{}, an area filter retained 9,994 boards from a random sample of 10,000.

\begin{table}[htbp]
\centering
\caption{Full marginal KS distances to D3-B test. The D3-B ref column measures variation between the two real-board splits. It is a reference value, not an achievable lower bound for a generator. SMD and THT denote surface-mount and through-hole pads, respectively.}
\label{tab:ks-full}
\small
\setlength{\tabcolsep}{7pt}
\begin{tabular}{lrrr}
\toprule
Statistic & D3-B ref & \cadetdata{} & D2 \\
\midrule
Board size, $\sqrt{WH}$ (mm) & 0.064 & 0.102 & 0.972 \\
Aspect ratio, $\ln(\mathrm{long}/\mathrm{short})$ & 0.170 & 0.124 & 0.606 \\
Net density (nets/cm$^2$) & 0.071 & 0.154 & 1.000 \\
Pad density (pads/cm$^2$) & 0.057 & 0.167 & 1.000 \\
Locality, minimum spanning tree (MST) edge length / board diagonal & 0.026 & 0.078 & 0.369 \\
Pads per net & 0.015 & 0.014 & 0.101 \\
SMD pad short side (mm) & 0.078 & 0.093 & 0.979 \\
SMD pad long side (mm) & 0.087 & 0.094 & 0.835 \\
THT pad size (mm) & 0.106 & 0.233 & 0.910 \\
THT drill diameter (mm) & 0.108 & 0.116 & 0.970 \\
Netclass clearance (mm) & 0.045 & 0.057 & 0.919 \\
Netclass track width (mm) & 0.059 & 0.031 & 0.599 \\
Netclasses per board & 0.028 & 0.108 & 0.645 \\
THT fraction of pads & 0.057 & 0.059 & 0.429 \\
Front-side fraction of SMD pads & 0.044 & 0.029 & 0.714 \\
Nets per board & 0.050 & 0.207 & 1.000 \\
Pads per board & 0.021 & 0.245 & 1.000 \\
Mean of all & 0.064 & 0.112 & 0.768 \\
\bottomrule
\end{tabular}
\end{table}

\cadetdata{} has smaller KS distances than D2 on every measured axis. Its distances range from 0.014 for pins per net to 0.245 for pads per board. Distances for aspect ratio, pins per net, track width, and the front-side fraction of SMD pads are below the corresponding D3-B ref--test references. These distances compare marginal distributions.

\begin{table}[htbp]
\centering
\caption{Selected proportions. Proportions are fractions, not percentages.}
\label{tab:joint-statistics}
\small
\setlength{\tabcolsep}{6pt}
\begin{tabular}{lrrrr}
\toprule
Statistic & D3-B test & D3-B ref & \cadetdata{} & D2 \\
\midrule
All-THT boards & 0.262 & 0.291 & 0.297 & 0.000 \\
All-front SMD & 0.615 & 0.620 & 0.621 & 0.005 \\
Two-pad nets & 0.610 & 0.600 & 0.616 & 0.601 \\
Three-pad nets & 0.194 & 0.195 & 0.193 & 0.199 \\
\bottomrule
\end{tabular}
\end{table}

\cref{tab:joint-statistics} provides complementary summaries. \cadetdata{} approximates several discrete proportions, including through-hole-only boards and two- and three-pad nets.

\subsection{Generation Details}
\label{app:generation-details}

\begin{table}[htbp]
\centering\small
\caption{D3-B ref sampling models before connection acceptance.}
\label{tab:generator-statistics}
\begin{tabular}{ll}
\toprule
Statistic & Sampling model \\
\midrule
Size $\sqrt{WH}$, net density & Lognormal \\
Log aspect ratio & Point mass and half-normal \\
Netclass sets, layer fractions, $P_i$ & Empirical resampling \\
Pad dimensions and drill & Empirical, conditional on type \\
Same-net MST edge length$/d$ & Lognormal locality \\
\bottomrule
\end{tabular}
\end{table}

\textbf{Sampling distributions.}\quad
Sampling statistics use the 142 D3-B ref boards. The fitted size variable is $\sqrt{WH}$, where $W,H$ are the board bounding-box dimensions in millimeters. Its lognormal distribution has median 39.95 mm and log standard deviation 0.458. Samples are clipped to $[10,105]$ mm. The log aspect ratio $\log(\max(W,H)/\min(W,H))$ has a point mass of 0.128 at zero. Otherwise, it follows a half-normal distribution with scale 0.655. The exponentiated draw is the long-to-short side ratio. It is capped at 4.3 and subsequently at $(\sqrt{WH}/12)^2$ before conversion to side lengths. Outlines are sampled from a fixed mixture of rectangles, rounded rectangles, polygons, and circles with probabilities 0.53, 0.25, 0.20, and 0.02. These shape probabilities are not fitted.

Net density $\rho_{\mathrm{net}}$, in nets/cm$^2$, follows a lognormal distribution with median 1.032 and log standard deviation 0.902. Size and density are sampled independently before connection acceptance. The calibrated scale $\alpha$ increases the net count to $\bar N$ through
\begin{equation}
\bar N=\operatorname{clip}_{[2,41]}\!\left(
\operatorname{round}\!\left(\alpha\rho_{\mathrm{net}}\frac{WH}{100}\right)\right),
\qquad P=\sum_{i=1}^{\bar N}P_i.
\label{eq:sampled-counts}
\end{equation}
Each $P_i$ is independently resampled from 2,506 observed net pad counts filtered to $[2,64]$. We sample one real board's complete netclass set and assign a class to each generated net using its observed class weights. The through-hole fraction and front-side surface-mount fraction are jointly resampled as one board-level pair. Pad type and surface-mount layer are then sampled using these fractions. Dimensions and drill diameters are resampled conditional on pad type, with width and height exchanged with probability 0.5.

For locality, a lognormal distribution is fitted to same-net MST edge lengths divided by the board diagonal. Its median is 0.1177 and log standard deviation is 0.90. This distribution proposes the second and subsequent pads around a uniformly selected accepted pad of the same net. The first pad of each net is proposed uniformly over the bounding box. Accepted distances need not retain the fitted distribution because site checks and connection search reject proposals.

With $d=\sqrt{W^2+H^2}$ and an accepted same-net anchor pad at $\bm{x}_{\mathrm{anc}}$, a subsequent pad $p$ is proposed at
\begin{equation}
\bm{x}_p^{\mathrm{prop}}=\bm{x}_{\mathrm{anc}}+du(\cos\varphi,\sin\varphi)^\top,
\quad u\sim\lognormal(\mu_{\mathrm{loc}},\sigma_{\mathrm{loc}}^2),\quad
\varphi\sim\mathcal U[0,2\pi).
\label{eq:pad-proposal}
\end{equation}
Here $\mu_{\mathrm{loc}}$ and $\sigma_{\mathrm{loc}}$ parameterize the lognormal locality distribution. The site is snapped to the placement grid before geometric checks.

\textbf{Construction gates and budgets.}\quad
Site sampling uses a 0.1 mm grid and at most 60 retries per site-sampling call. The generator allows up to 30 connection attempts for a pad. The first pad is provisional, and the first pair is committed only after a successful connection. Failure to find a site or exhaustion of connection attempts stops the current net, including when its first or second pad fails. A net with fewer than two committed pads is discarded. A 1,500 s construction limit is checked after each net and does not include finalization.

TryConnect loads the proposed board while retaining accepted routing. It rejects initial geometric DRVs and a new pad already connected without routing, selects its net, and searches for a connection. A post-connection check rejects remaining required connections and strict DRC failures, including dangling tracks. Successful proposals retain their pads, routing, and actions. Failed proposals leave the retained construction unchanged. Consequently, connectivity and zero evaluated geometric DRVs hold at accepted extensions, not necessarily at provisional sites or intermediate search states.

\textbf{Connection search.}\quad
Each queue entry $(\tau_p,a)$ stores the engine checkpoint, wrapper snapshot, observation, and routing head reached by $\tau_p$. Candidate actions comprise direct requests to up to three nearby same-layer targets, eight directional requests at each of four scales (0.2, 1, 5, and 25 mm), and a via to another layer. Diagonal requests multiply these scales by $\sqrt{2}$, so 25 mm is not a maximum hop distance. Each recorded hop pairs route initialization with a line or via operation. During search, initialization is executed only when the routing session is closed. The count $|\tau_p|$ in Equation~\eqref{eq:search-cost} counts the actions in $\tau_p$, not wirelength.

The construction cost in Equation~\eqref{eq:search-cost} uses walkaround mode.
 During final repair, candidates with $\bm{x}_a$ on $S_i$ also use push-and-shove and mark-obstacles modes. Their costs are

\begin{equation}
f(\tau_p,a)=2(|\tau_p|+1)+\delta_{\mathrm{mode}}(a),\qquad
\delta_{\mathrm{mode}}(a)=
\begin{cases}
0, & \text{walkaround},\\
0.5, & \text{push-and-shove},\\
1, & \text{mark-obstacles}.
\end{cases}
\label{eq:repair-search-cost}
\end{equation}
Other candidates retain Equation~\eqref{eq:search-cost} and use walkaround mode. Ties are resolved in insertion order. The heuristic is computed once per connection using multi-source Dijkstra on an eight-connected 0.2 mm grid. The grid uses optimistic obstacles and a layer-change cost equivalent to 30 mm. If the grid field is unavailable, the search uses a Euclidean fallback. The grid guides cost estimates, detour checks, and expansion-budget reduction.
 It does not determine legality.

The requested endpoint $\bm{x}_a$ in Equation~\eqref{eq:search-cost} comprises the requested coordinates and the layer after the requested operation. For a via this is a different layer. When a candidate action is selected, the search restores its parent checkpoint if necessary and executes the operation in the engine. Duplicate heads are compared at 0.01 mm precision. Candidates require an effective valid action, no increase in evaluated geometric DRVs, and no dangling-track overshoot. The initial pass also limits detours by $\Delta\ell\le1.3\Delta h+1$ mm, where $\Delta h=\max(0,\min(h_{\mathrm{before}},10^6)-\min(h_{\mathrm{after}},10^6))$. The heuristic values are evaluated at the current and next heads, with $h_{\mathrm{after}}=0$ upon connection. A failed connection is retried without this cap.

Search stops when it finds a connection, exhausts the queue, or reaches its expansion budget. Construction permits 60 expansions and final repair permits 200. A grid-unreachable target reduces the budget to at most 12. A connecting prefix is replayed linearly from its root before it is returned. The heuristic is not established to be admissible or consistent for the implemented hop cost.
The finite search does not guarantee a shortest path. Search-time geometric DRC filtering excludes connectivity-related reports and dangling tracks. The post-connection filter includes dangling tracks and excludes only missing connections, which are checked separately.

\begin{algorithm}[t]
\small
\caption{Finalization: Replay, Repair, and Pruning}
\label{alg:finalize}
\algtext*{EndFor}
\algtext*{EndIf}
\begin{algorithmic}[1]
\Require Unrouted instance $s_0$, action sequences $\tau_p\in\mathcal D$ in recorded order
\For{$k=1,\ldots,16$}
    \State $s\gets s_0$
    \For{each $\tau_p\in\mathcal D$ in recorded order}
        \If{replay of $\tau_p$ from $s$ fails}
            \State $\tau_p\gets\mathrm{TryConnect}(p,S_i)$ from $s$ with the cost in Equation~\eqref{eq:repair-search-cost} \Comment{repair}
            \If{$\tau_p=\mathrm{FAIL}$}
                \State Prune $p$ and continue with the next $k$
            \EndIf
        \EndIf
        \State $s\gets F(s,\tau_p)$
    \EndFor
    \If{fresh-engine replay of $\mathcal D$ from $s_0$ executes every action, connects all nets, and yields zero DRVs}
        \State \Return $(s_0,\mathcal D)$ \Comment{accept}
    \EndIf
    \If{the first divergent action lies in no $\tau_p$}
        \State \Return FAIL
    \EndIf
    \State Prune $p$ if its $\tau_p$ holds the first divergent action for the third time
\EndFor
\State \Return FAIL
\end{algorithmic}
\end{algorithm}

\textbf{Finalization and replay.}\quad
The generator removes construction routing and assembles recorded connection actions with net-selection and net-ending operations. Algorithm~\ref{alg:finalize} replays this sequence on the completed unrouted instance. Replay of $\tau_p$ fails when an action fails, unconnected links do not decrease, or strict-check DRVs remain. Pruning removes pad $p$ and its sequence $\tau_p$, and it removes nets with fewer than two remaining pads.

Final acceptance requires plain replay on a fresh engine server without search or checkpoint restoration. Its terminal DRV count omits the intermediate filters. The exact unrouted board used for this check is retained with the action sequence. Construction geometry need not be reproduced exactly during final replay. Recorded operations specify action type, target coordinates and layers, and routing mode. Replay resolves coordinates against live candidates and permits a 0.8 mm snapping fallback for via targets.

\textbf{Realized counts and calibration.}\quad
The mathematical net index persists through pruning. A removed net has $\hat P_i=0$, giving
\begin{equation}
\hat P=\sum_{i=1}^{\bar N}\hat P_i,
\qquad
\hat N=\sum_{i=1}^{\bar N}\bm{1}[\hat P_i\ge2].
\label{eq:realized-counts}
\end{equation}
These indices are mathematical bookkeeping. Stored net codes can be renumbered, and the saved witness does not preserve a direct original-to-final net-code mapping.

For the accepted pilot boards $\mathcal S_r$ at scale $\alpha_r$, the pilot update is
\begin{equation}
\alpha_{r+1}=\alpha_r\frac{m_{\mathrm{target}}}{m_r},
\qquad
m_r=\operatorname{median}_{s\in\mathcal S_r}\frac{100\hat P(s)}{W(s)H(s)},
\label{eq:density-calibration}
\end{equation}
where $m_r$ is the median realized pad density in pads/cm$^2$.
The factor 100 converts area in mm$^2$ to cm$^2$. The D3-B ref target is $m_{\mathrm{target}}=3.628$ pads/cm$^2$. The sampled quantity multiplied by $\alpha$ remains net density. Each pilot attempts 3,000 boards with fresh seeds. Pilots stop when $m_{\mathrm{target}}/m_r$ lies within 2\% of one or after four iterations. The pilots used scales 1.0000, 1.3324, 1.5191, and 1.6074. Its final pilot ratio was 1.0268, outside the tolerance. Production used the subsequent update, 1.6505, without another pilot. The procedure therefore stopped at its iteration limit without meeting the tolerance. Median correction does not establish agreement of distribution tails or dependencies between statistics.

\subsection{Acceptance Rate and Runtime}
\label{app:generation-cost}

We report costs for one server with two AMD EPYC 7453 CPUs (56 cores) running 80 worker processes.
The times below follow from the throughput measured on each of 17 such servers, which generated boards in parallel without coordination.
Generation accepts about 920 boards per hour on one server.
Production attempted 106,000 boards and accepted 105,066 (99.1\%), which takes about 115 hours on one server.
The remaining 934 boards failed finalization or raised an error.
Most rejections occur at the pad level and remove only the rejected pad, so few boards are discarded.
The median wall-clock time per accepted board was 106.5 seconds, and the 90th percentile was 845 seconds. The 90th percentiles were 783 seconds for construction, 741 seconds for TryConnect search within it, and 59 seconds for replay with repair and pruning.
The four calibration pilots take about 13 hours, and the independent replay of the first 100,000 accepted demonstrations takes about 18 hours.
The full pipeline therefore takes about 146 hours, or six days, on one server.
Every replayed demonstration executed all actions, connected all nets, and yielded zero DRVs.

\section{Training Details}
\label{app:training-details}

\textbf{Architecture and PPO.}\quad
All policies used the Transformer policy of PCBWorld~\citep{pcbworld2026}, enlarged from four to six layers, width 128 to 256, four to eight attention heads, and feed-forward width 512 to 1,024.
For a fair comparison, we retrained our reimplementation of PCBWorld, PPO (D2), at this enlarged size, and it exceeds the original policy in CP on D3-A.
\cref{tab:ppo-hyperparameters} lists the architecture and PPO hyperparameters shared by all runs.
\rev{Every learned policy was trained with four training seeds, 42, 43, 44, and 45.
For \ours{}, each seed ran BC and then RL fine-tuning. For the PPO baselines, each seed ran scratch PPO.
The selected checkpoint of each seed was evaluated with four runs per board.
Reported values are means over the four seeds, and standard deviations and confidence intervals are also over seeds (Appendix~\ref{app:full-results}).}

\begin{table}[htbp]
\centering
\caption{Architecture and PPO hyperparameters shared by all policies. \cref{tab:training-settings} lists the settings that differ between runs.}
\label{tab:ppo-hyperparameters}
\small
\begin{tabular}{ll}
\toprule
Hyperparameter & Value \\
\midrule
\multicolumn{2}{l}{\emph{Architecture}} \\
Transformer layers & 6 \\
Embedding width & 256 \\
Attention heads & 8 \\
Feed-forward width & 1,024 \\
Parameters & 6.79M \\
\midrule
\multicolumn{2}{l}{\emph{PPO}} \\
Parallel environments & 64 \\
Rollout steps per environment & 1,024 \\
Update epochs & 4 \\
Batch size & 128 \\
Learning rate & $2\times10^{-5}$ \\
Clipping parameter & 0.1 \\
Discount factor & 0.995 \\
Generalized advantage estimation (GAE) parameter & 0.95 \\
Reward normalization & Enabled \\
Advantage normalization & Enabled \\
Entropy coefficient & 0 \\
\rev{Training seeds} & \rev{42, 43, 44, 45} \\
\bottomrule
\end{tabular}
\end{table}

The policies shared the configured reward rule, DRC observations, multi-resolution directional candidates, off-board masking, and connectivity filtering.
No run overrode the reward coefficients.
PPO (D2) used the same Transformer dimensions but omitted obstacle, shape, and per-net constraint features.
It represented outlines with polygonal segments and encoded edges through their endpoints.
The \cadetdata{} policies additionally encoded an edge midpoint to represent native arcs.
PPO (D3-B ref) used the same architecture and full observation settings as scratch training on \cadetdata{}.

\textbf{Behavior cloning.}\quad
BC trained the same policy from random initialization by maximizing the log-likelihood of the recorded actions in the 100,000 \cadetdata{} training demonstrations.
It used the action masks of the PPO update, and replay of each demonstration in the engine reconstructed its states.
Optimization used AdamW with $(\beta_1,\beta_2)=(0.9,0.999)$, $\epsilon=10^{-5}$, and weight decay $10^{-4}$.
The learning rate warmed up linearly over 2,000 steps to $4\times10^{-4}$ and then decayed by a cosine schedule to 5\% of this peak.
Training ran for 350,000 steps with 512 state--action pairs per step, gradient-norm clipping at 1.0, \rev{and bf16 autocast.
BC ran once per training seed.}
Validation negative log-likelihood (NLL) was computed every 2,000 steps on 2,048 state--action pairs from 128 held-out \cadetdata{} validation demonstrations.
\rev{For each seed, among the checkpoints saved every 35,000 steps, the one with the lowest validation NLL initialized RL fine-tuning.
These were steps 350,000, 315,000, 280,000, and 350,000 for seeds 42 to 45, respectively.}
No real board entered BC training or checkpoint selection.

\textbf{Fine-tuning and scratch PPO.}\quad
\ours{} fine-tuned the BC policy for 200 PPO iterations with the KL penalty of Equation~\eqref{eq:ppo-kl} and $\beta=0.05$.
Scratch PPO trained from random initialization for 500 iterations without the KL penalty.
BC does not train the critic, so fine-tuning began with a critic warm-up.
Critic warm-up froze the actor for at least ten iterations.
It stopped when critic explained variance reached 0.9 for three consecutive iterations, with a 50-iteration cap.
With critic detachment, the critic loss does not update the encoder shared with the actor.
Scratch runs used neither setting.
\cref{tab:training-settings} summarizes the differing settings.

\begin{table}[htbp]
\centering
\caption{Configured policy differences. Scratch denotes PPO on \cadetdata{}, its variant without TryConnect, or D3-B ref. Warm-up values denote configured minimum and maximum lengths, not measured durations.}
\label{tab:training-settings}
\small
\setlength{\tabcolsep}{5pt}
\begin{tabular}{lccc}
\toprule
Setting & \ours{} & Scratch & PPO (D2) \\
\midrule
Initialization & \rev{BC} & Random & Random \\
Obstacle / shape / net constraints & Enabled & Enabled & Disabled \\
Outline representation & Native arcs & Native arcs & Polygonal \\
Edge midpoint encoding & Enabled & Enabled & Disabled \\
KL coefficient & 0.05 & 0 & 0 \\
Critic warm-up (iterations) & 10--50 & 0 & 0 \\
Critic detachment & Enabled & Disabled & Disabled \\
\bottomrule
\end{tabular}
\end{table}

\section{Evaluation Metrics}
\label{app:metrics}

All metrics are computed from the final board file of a rollout with the post-hoc \kicad{} DRC of \S\ref{sec:setup}.
Let $U(s)$ be the number of ratsnest lines, i.e., unrouted connections, that \kicad{} reports in $s$.
Let $n_{\mathrm{err}}(s)$ be the number of error-severity DRVs, which include unrouted connections, and let $\ell(s)$ and $n_{\mathrm{via}}(s)$ be the wirelength and via count of \S\ref{sec:routing-mdp}.
For a rollout from $s_0$ that ends in $s_T$, the per-board metrics are
\begin{align}
\mathrm{CP}&=\bm{1}\bigl[n_{\mathrm{err}}(s_T)=0\bigr], &
\mathrm{Rout}&=1-\frac{U(s_T)}{U(s_0)}, &
\Delta\Phi&=\Phi(s_T)-\Phi(s_0),
\label{eq:main-metrics}\\
\mathrm{WL}&=\ell(s_T), &
\mathrm{Vias}&=n_{\mathrm{via}}(s_T), &
\mathrm{DRVs}&=n_{\mathrm{err}}(s_T).
\label{eq:layout-metrics}
\end{align}
A rollout that produced no board file counts as unrouted with $s_T=s_0$.
A run scores a metric $m$ by its mean $\frac{1}{|\mathcal E|}\sum_{b\in\mathcal E}m_b$ over the boards $b$ of an evaluation set $\mathcal E$.
Tables report the mean of this score over runs \rev{and training seeds}, and \cref{tab:full-results} adds its standard deviation.

\textbf{Selection @$k$.}\quad
In a run of a method @$k$, board $b$ has $k$ rollouts with metrics $\mathrm{CP}_b^{(j)}$ and $\Delta\Phi_b^{(j)}$ for $j=1,\ldots,k$.
The run keeps the rollout
\begin{equation}
j_b^\star=\operatorname*{arg\,max}_{j\in\{1,\ldots,k\}}\bigl(\mathrm{CP}_b^{(j)},\,\Delta\Phi_b^{(j)}\bigr),
\label{eq:at-k}
\end{equation}
where the maximum compares pairs lexicographically, and all metrics of board $b$ except Time come from rollout $j_b^\star$.
\ours{} ran 12 rollouts per board.
\ours{} @1 uses the first four as four runs, and \ours{} @3 splits all 12 into four runs of three consecutive rollouts.

\textbf{Time.}\quad
Let $t_b^{(j)}$ be the wall-clock time of rollout $j$ on board $b$, measured with one board at a time on the server of \S\ref{sec:setup}.
Time is the board mean of $\mathrm{Time}_b=\sum_{j=1}^{k}t_b^{(j)}$, so @$k$ pays for all $k$ rollouts.
For learned policies, $t_b^{(j)}$ covers board loading, environment reset, the episode, and board saving, and excludes model loading.
For LLM agents, it covers the routing episode.
For rule-based routers, it covers router startup, input parsing, routing, and output conversion, and a run that reaches the 1,800-second limit counts as 1,800 seconds.
Time excludes the post-hoc DRC, including the DRC used for @$k$ selection.

\section{Routing Environment}
\label{app:environment}

This section summarizes the parts of the environment in \S\ref{sec:routing-mdp} that the paper uses. \citet{pcbworld2026} give the full specification.

\textbf{Episode.}\quad An episode starts from the unrouted instance $s_0$. It ends when all required connections are complete or at the step limit. After each action, the engine updates connectivity and reports DRVs, which enter the next state and the reward. The state also holds the routing head, which records the active net, the current position and layer, and the routing phase. At each step, the environment masks actions that the current phase does not allow, such as a line operation before a route starts.

\textbf{Actions.}\quad We use five engine-level actions.
\begin{itemize}
\item \texttt{net\_select}($i$) makes net $i$ the active net and applies its rule vector $\bm{r}_i$ in the engine.
\item \texttt{start\_route}($\bm{x}$, $l$) places the routing head at location $\bm{x}$ on layer $l$, such as a pad or the end of existing copper of the active net.
\item \texttt{make\_line}($\bm{x}_a$, $m$) routes from the routing head toward the requested endpoint $\bm{x}_a$ in routing mode $m$. The engine constructs the track segments and moves the routing head to their end.
\item \texttt{make\_via}($\bm{x}_a$, $m$) routes to $\bm{x}_a$ as \texttt{make\_line} does, places a via there, and moves the routing head to a different layer.
\item \texttt{net\_end} ends routing of the active net and returns to net selection.
\end{itemize}
The routing mode $m$ selects how the engine treats blocking copper of other nets. Mark-obstacles mode stops at it, walkaround mode detours around it, and push-and-shove mode moves it aside. \pcbworld{} also provides a \texttt{finish} action that automatically routes to the nearest unconnected pad. We disable it in all experiments, so every connection comes from the five actions above.
\textbf{Action parameterization.}\quad LLM agents write the action name and its numeric arguments as text (Appendix~\ref{app:llm-prompt}). RL policies and TryConnect instead choose from the candidate action set $\mathcal A(s)$. Each candidate combines an action type, an endpoint, and a routing mode. Endpoints include pad and copper locations and directional points in eight directions at offsets of 0.2, 1, 5, and 25~mm per axis from the routing head. The environment masks endpoints outside the board outline.

\raggedbottom
\section{Demonstration Example}
\label{app:demo-example}

This section shows the first 14 actions of a \cadetdata{} validation demonstration, which route two nets from $s_0$. Each net begins with \texttt{net\_select} and ends with \texttt{net\_end}. Between them, each hop pairs \texttt{start\_route} with \texttt{make\_line} or \texttt{make\_via} (Appendices~\ref{app:environment} and~\ref{app:generation-details}). Each panel shows the state $s_t$ and $a_t$ in cyan. For actions with a location, white dots mark the candidates of $a_t$ in $\mathcal A(s_t)$.

\begin{center}
\captionsetup{type=figure}
\begin{subfigure}[t]{0.46\linewidth}
\includegraphics[width=\linewidth,trim=12 12 12 12,clip]{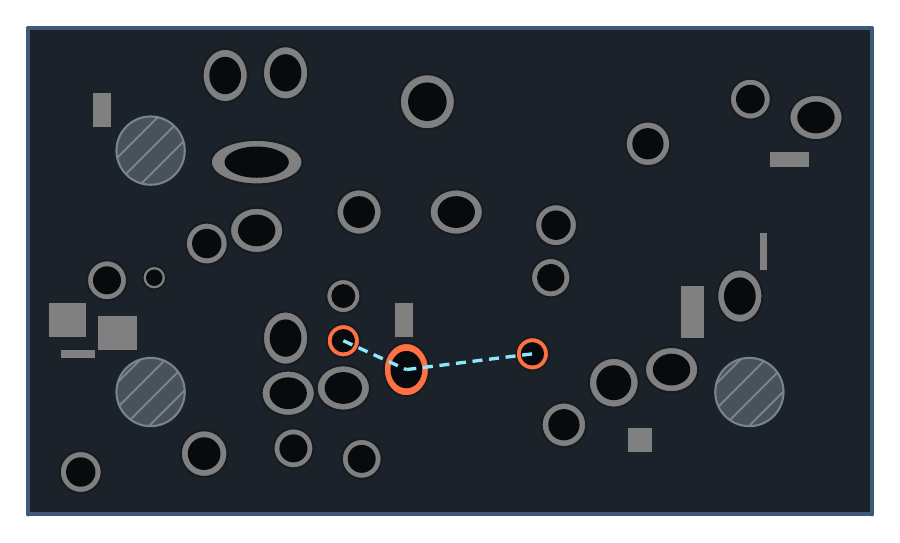}
\caption{$t=0$, \texttt{net\_select}(net 1)}
\end{subfigure}\hfill
\begin{subfigure}[t]{0.46\linewidth}
\includegraphics[width=\linewidth,trim=12 12 12 12,clip]{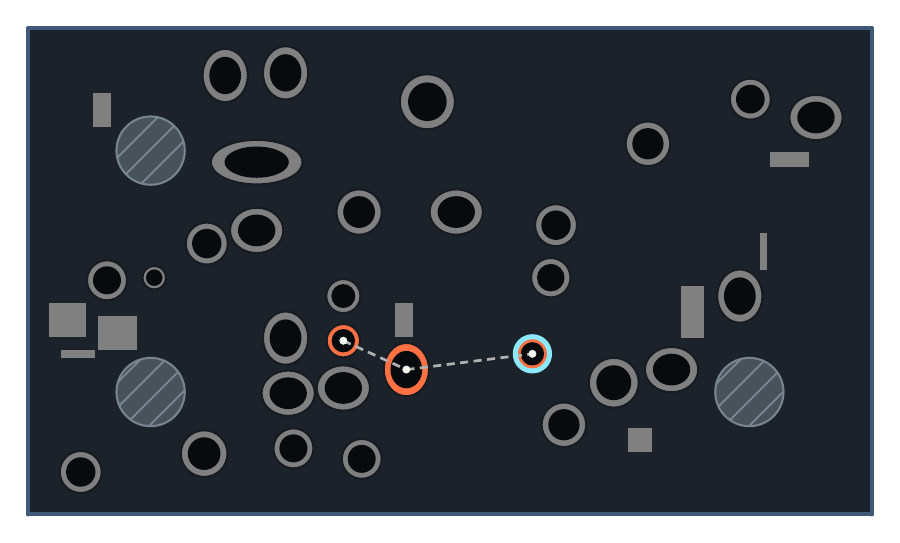}
\caption{$t=1$, \texttt{start\_route}(19.2, 12.4, F.Cu)}
\end{subfigure}
\par\vspace{4pt}
\begin{subfigure}[t]{0.46\linewidth}
\includegraphics[width=\linewidth,trim=12 12 12 12,clip]{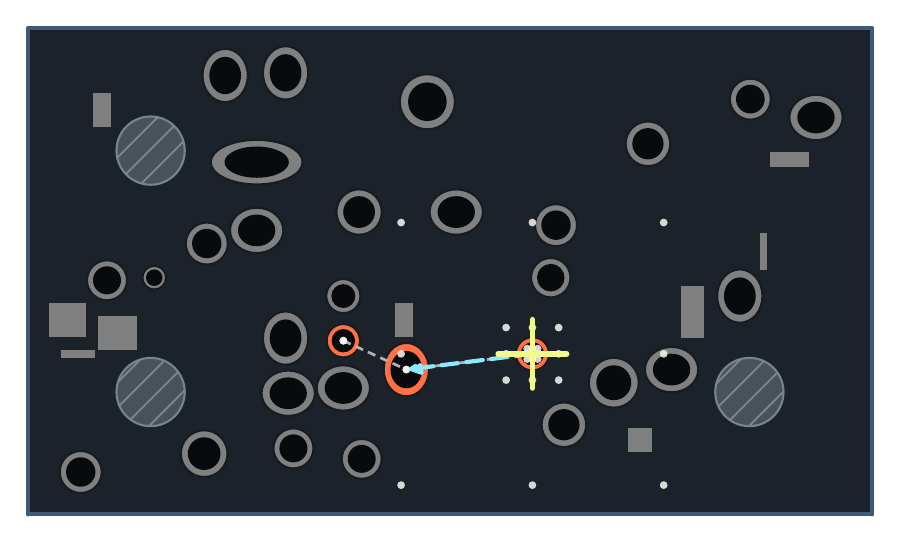}
\caption{$t=2$, \texttt{make\_line}(14.4, 13.0, walkaround)}
\end{subfigure}\hfill
\begin{subfigure}[t]{0.46\linewidth}
\includegraphics[width=\linewidth,trim=12 12 12 12,clip]{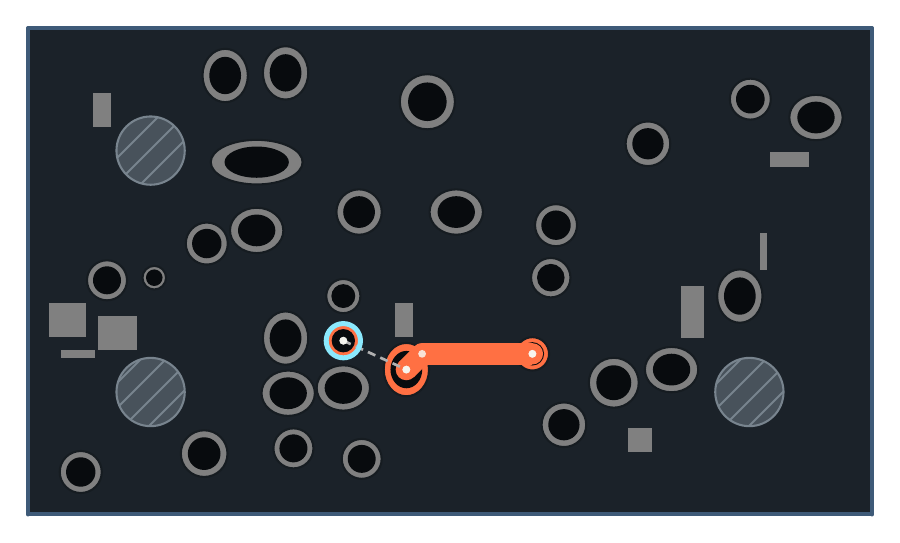}
\caption{$t=3$, \texttt{start\_route}(12.0, 11.9, F.Cu)}
\end{subfigure}
\caption{Net~1 at $t=0$ to 3. Each panel shows $s_t$ and, in cyan, $a_t$ (coordinates in mm). Orange marks the current net and gray the other nets. White dots mark candidates, and the yellow crosshair marks the routing head. A white dashed center line marks a back-layer track, and a gray dashed line an unrouted connection. In cyan, a dashed line marks \texttt{net\_select}, a ring \texttt{start\_route}, an arrow \texttt{make\_line}, and an arrow ending in a filled dot \texttt{make\_via}.}
\label{fig:demo-1}
\end{center}

In \cref{fig:demo-1}, \texttt{net\_select} chooses net~1 at $t=0$. Its three orange pads are to be connected, and the cyan dashed line shows their unrouted connections. At $t=1$, \texttt{start\_route} chooses a start point among the pad centers (cyan ring) and places the routing head there (yellow crosshair at $t=2$). At $t=2$, the white dots are the requested endpoints of \texttt{make\_line} from the routing head, including the center of another pad. The chosen \texttt{make\_line} ends at that pad center, and the engine routes a track to it. Each recorded hop begins with \texttt{start\_route}. At $t=3$, it chooses among the pads and track ends and places the routing head at the unconnected pad.

\begin{center}
\captionsetup{type=figure}
\begin{subfigure}[t]{0.46\linewidth}
\includegraphics[width=\linewidth,trim=12 12 12 12,clip]{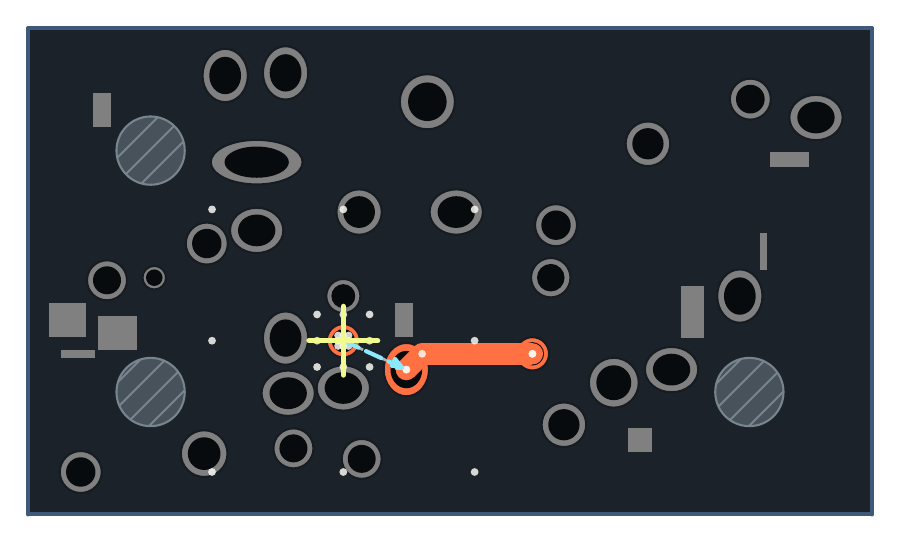}
\caption{$t=4$, \texttt{make\_line}(14.4, 13.0, walkaround)}
\end{subfigure}\hfill
\begin{subfigure}[t]{0.46\linewidth}
\includegraphics[width=\linewidth,trim=12 12 12 12,clip]{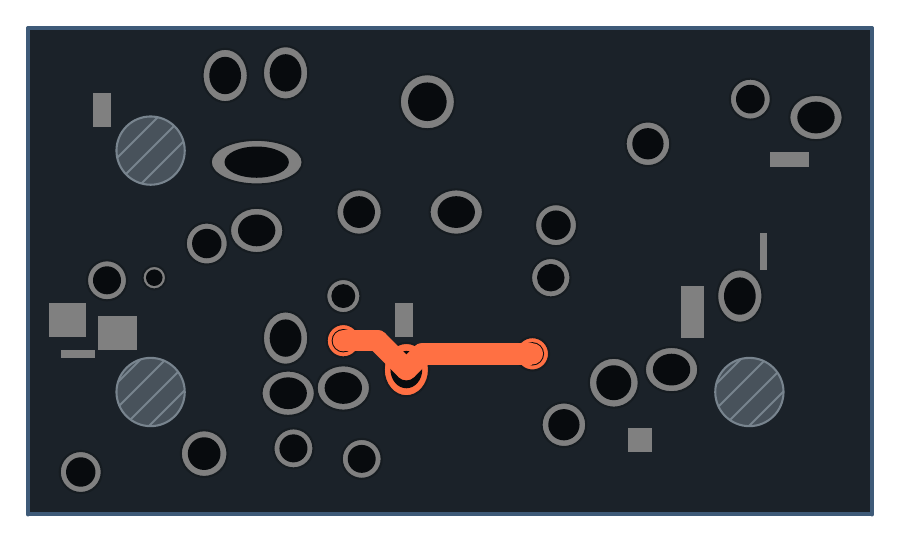}
\caption{$t=5$, \texttt{net\_end}()}
\end{subfigure}
\caption{Net~1 at $t=4$ and 5. Markings follow \cref{fig:demo-1}.}
\label{fig:demo-2}
\end{center}

In \cref{fig:demo-2}, \texttt{make\_line} connects the remaining pad at $t=4$, and \texttt{net\_end} finishes net~1 at $t=5$.

\begin{center}
\captionsetup{type=figure}
\begin{subfigure}[t]{0.46\linewidth}
\includegraphics[width=\linewidth,trim=12 12 12 12,clip]{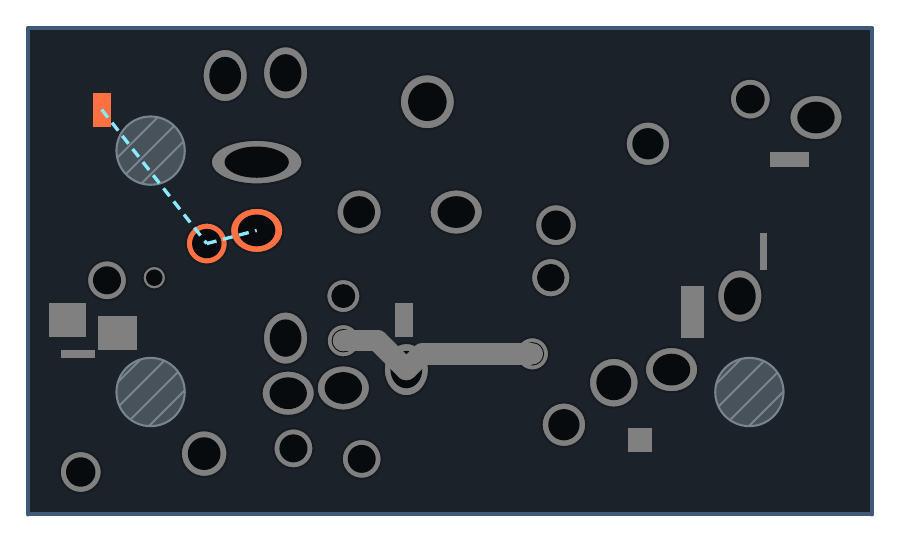}
\caption{$t=6$, \texttt{net\_select}(net 2)}
\end{subfigure}\hfill
\begin{subfigure}[t]{0.46\linewidth}
\includegraphics[width=\linewidth,trim=12 12 12 12,clip]{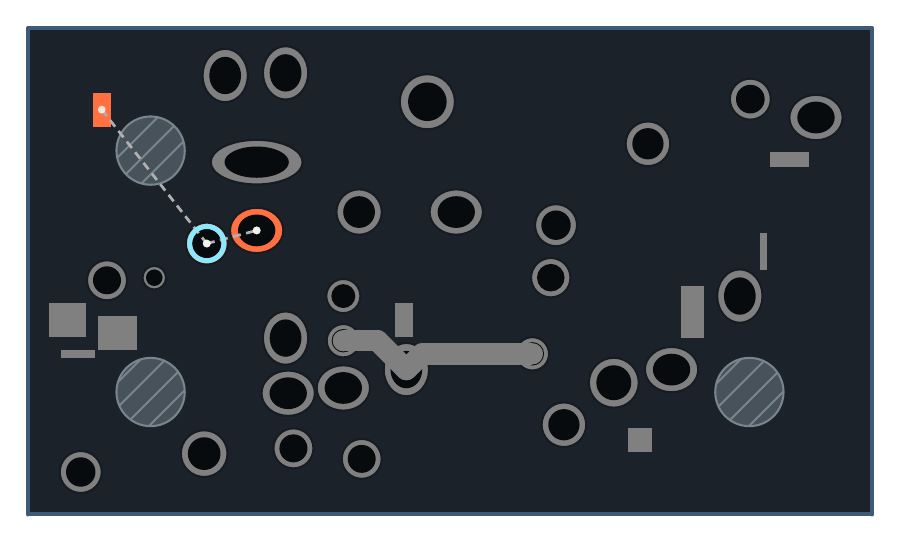}
\caption{$t=7$, \texttt{start\_route}(6.8, 8.2, F.Cu)}
\end{subfigure}
\par\vspace{4pt}
\begin{subfigure}[t]{0.46\linewidth}
\includegraphics[width=\linewidth,trim=12 12 12 12,clip]{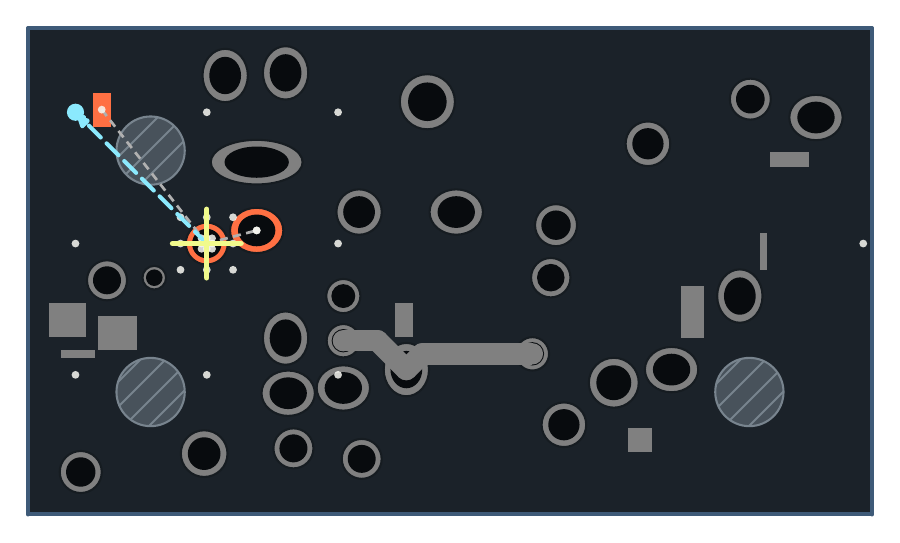}
\caption{$t=8$, \texttt{make\_via}(1.8, 3.2, walkaround)}
\end{subfigure}\hfill
\begin{subfigure}[t]{0.46\linewidth}
\includegraphics[width=\linewidth,trim=12 12 12 12,clip]{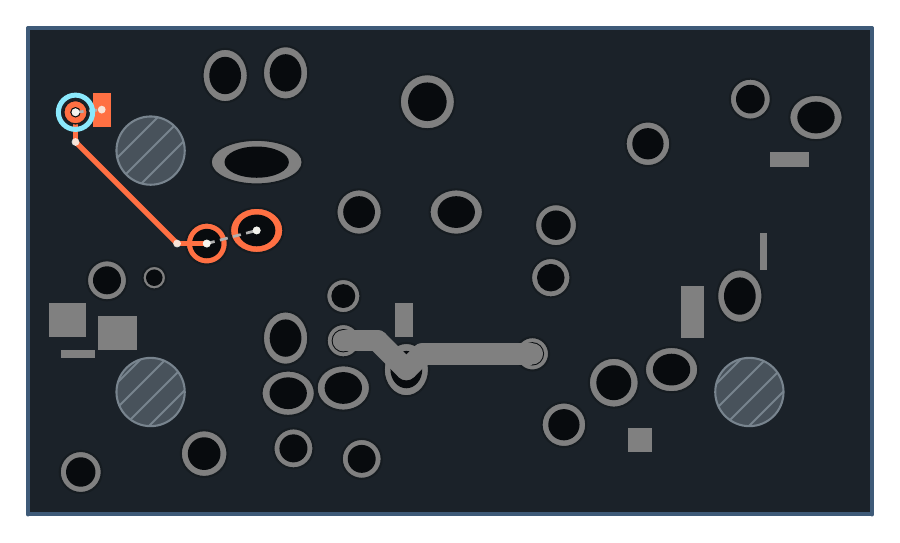}
\caption{$t=9$, \texttt{start\_route}(1.8, 3.2, B.Cu)}
\end{subfigure}
\caption{Net~2 at $t=6$ to 9. Markings follow \cref{fig:demo-1}.}
\label{fig:demo-3}
\end{center}

In \cref{fig:demo-3}, \texttt{net\_select} chooses net~2 at $t=6$, and net~1 turns gray. At $t=7$, \texttt{start\_route} starts at the round pad. At $t=8$, \texttt{make\_via} requests a free point rather than a pad center (filled cyan dot). The engine routes a front-layer track there and places a via. This single action creates three track segments. At $t=9$, \texttt{start\_route} starts from the via on the back layer.

\begin{center}
\captionsetup{type=figure}
\begin{subfigure}[t]{0.46\linewidth}
\includegraphics[width=\linewidth,trim=12 12 12 12,clip]{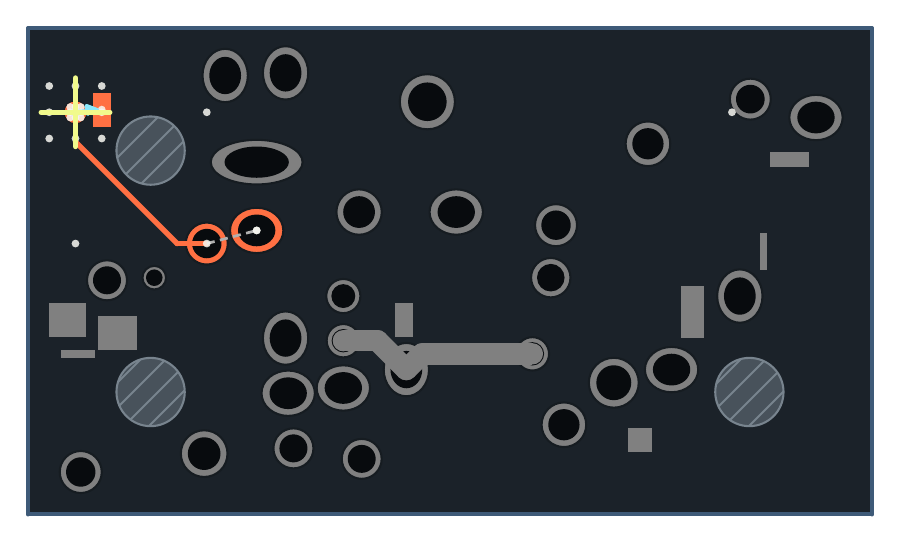}
\caption{$t=10$, \texttt{make\_line}(2.8, 3.1, walkaround)}
\end{subfigure}\hfill
\begin{subfigure}[t]{0.46\linewidth}
\includegraphics[width=\linewidth,trim=12 12 12 12,clip]{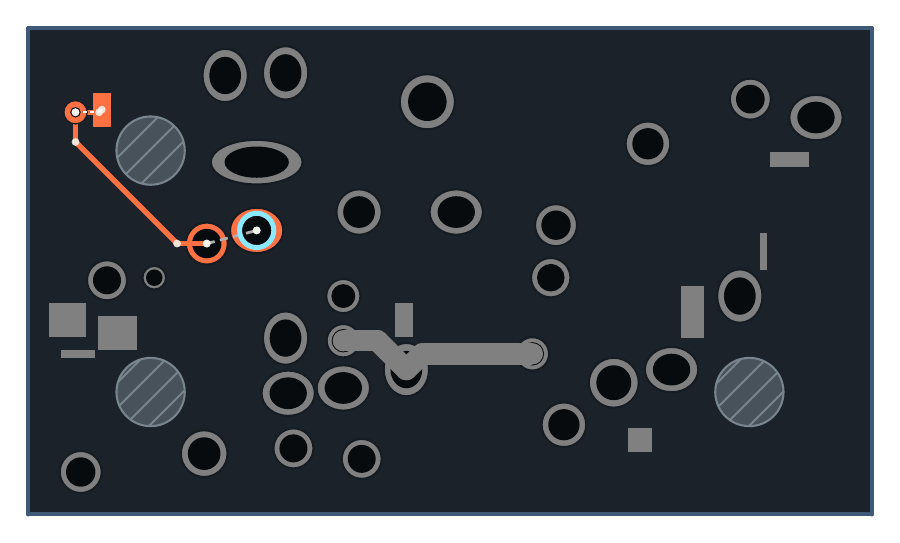}
\caption{$t=11$, \texttt{start\_route}(8.7, 7.7, F.Cu)}
\end{subfigure}
\par\vspace{4pt}
\begin{subfigure}[t]{0.46\linewidth}
\includegraphics[width=\linewidth,trim=12 12 12 12,clip]{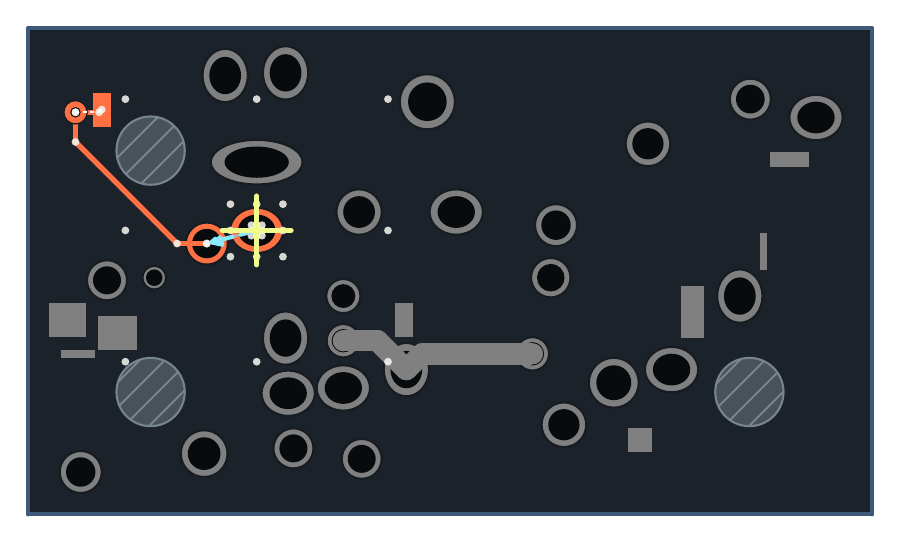}
\caption{$t=12$, \texttt{make\_line}(6.8, 8.2, walkaround)}
\end{subfigure}\hfill
\begin{subfigure}[t]{0.46\linewidth}
\includegraphics[width=\linewidth,trim=12 12 12 12,clip]{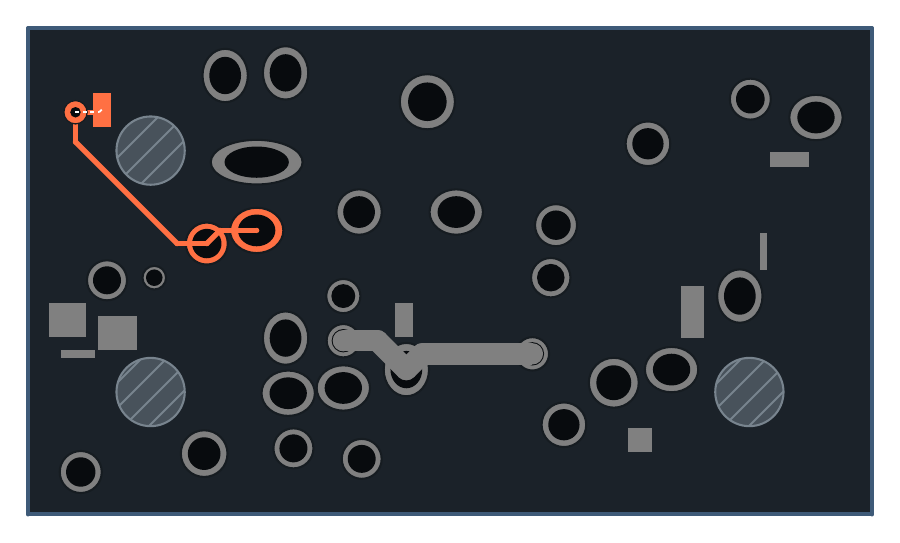}
\caption{$t=13$, \texttt{net\_end}()}
\end{subfigure}
\caption{Net~2 at $t=10$ to 13. Markings follow \cref{fig:demo-1}.}
\label{fig:demo-4}
\end{center}

In \cref{fig:demo-4}, a short back-layer \texttt{make\_line} reaches the surface-mount pad on that layer at $t=10$. At $t=11$, \texttt{start\_route} chooses the remaining unconnected pad, and \texttt{make\_line} connects it to the round pad at $t=12$. At $t=13$, \texttt{net\_end} finishes net~2.

\flushbottom
\section{Reward Potential}
\label{app:reward}

This section specifies the potential $\Phi$ of \S\ref{sec:routing-mdp}.
Let $\mathcal R$ be the set of nets with at least two pads, and let $U(s)$ be as in Appendix~\ref{app:metrics}. For net $i$ in state $s$, let $g_i(s)$ be the number of connected components formed by its pads and copper.
Let $N_{\mathrm{con}}(s)=|\{i\in\mathcal R:g_i(s)=1\}|$ be the number of fully connected nets.
Unlike $n_{\mathrm{err}}$, the DRVs in $\Phi$ also include dangling tracks, dangling vias, and net conflicts.
Let $n_{\mathrm{drv}}(s)$ be their total count, $\mathcal V(s)$ the set of nets with at least one of them, and $v_k(s)$ their count on net $k$.
The set $\mathcal V^+(s)$ adds to $\mathcal V(s)$ one entry $k=0$ that collects DRVs attributed to no net, when such DRVs exist.
The potential is
\begin{align}
\Phi_{\mathrm{con}}(s)&=-U(s)+N_{\mathrm{con}}(s)+|\mathcal R|-|\mathcal V(s)|+b_{\mathrm{done}}\bigl(\bm{1}[U(s)=0]+\bm{1}[n_{\mathrm{drv}}(s)=0]\bigr),
\label{eq:phi-con}\\
f_{\mathrm{drv}}(s)&=3\ln\!\Bigl(1+\frac{|\mathcal V^+(s)|}{2}\Bigr)+\sum_{k\in\mathcal V^+(s)}\ln\!\Bigl(1+\frac{v_k(s)}{2}\Bigr),
\label{eq:f-drv}
\end{align}
where $b_{\mathrm{done}}=\frac12\sum_{i\in\mathcal R}g_i(s_0)$ is computed at each reset.
The quality weights are $\lambda_w=0.5/L$ with $L=\max(W,H)$ and $\ell$ in millimeters, and $\lambda_v=0.2$.
The reward is
\begin{equation}
r_t=\Phi(s_{t+1})-\Phi(s_t)-0.01\,\bm{1}[a_t\text{ is invalid}],
\label{eq:reward-full}
\end{equation}
where an action is invalid if the engine rejects it or it has no effect.
No terminal reward is added.
The penalty enters only $r_t$, so $\Delta\Phi$ in Appendix~\ref{app:metrics} excludes it.

\section{Full Evaluation Results}
\label{app:full-results}

\cref{tab:full-results} reports all evaluated methods with standard deviations and three additional layout metrics, WL, Vias, and DRVs of Equation~\eqref{eq:layout-metrics}.
\rev{The mean and standard deviation are over four training seeds for the learned methods and over four runs for FreeRouting.
With four seeds or runs, the 95\% confidence interval (CI) of a mean is the mean $\pm$ 1.59 standard deviations, from the $t$ distribution. \cref{tab:cp-ci} gives these CIs for CP.}
Unrouted connections contribute no tracks, so lower routability tends to shorten wirelength.

\begin{table}[H]
\centering
\caption{Full evaluation results as mean $\pm$ standard deviation \rev{over four training seeds for the learned methods and over four runs for FreeRouting}. KRT and the LLM agents ran once and have no standard deviation. All methods except \ours{} @3 use one rollout per run. CP and Rout. are in percent, WL is in millimeters, and Time is in seconds per board.}
\label{tab:full-results}
\footnotesize
\setlength{\tabcolsep}{2pt}
\newcommand{\sd}[1]{{\scriptsize$\pm$#1}}
\begin{tabular}{lrrrrrrr}
\toprule
Method & CP & $\Delta\Phi$ & Rout. & WL & Vias & DRVs & Time \\
\midrule
\multicolumn{8}{l}{\textit{D3-B test (141)}} \\
Reference & 100.0 & 144.41 & 100.0 & 575 & 6.6 & 0.00 & -- \\
FreeRouting @1 & 65.6\sd{4.1} & 119.33\sd{3.22} & 96.8\sd{0.1} & 527\sd{5} & 4.1\sd{0.2} & 1.14\sd{0.22} & 45.6\sd{6.0} \\
KRT @1 & 27.7 & 84.71 & 78.3 & 463 & 14.7 & 8.23 & 2.7 \\
GPT-6 Astra @1 & 56.0 & 104.20 & 86.0 & 466 & 7.5 & 6.81 & 491.7 \\
GPT-6 Luna @1 & 5.0 & 37.72 & 47.9 & 203 & 2.0 & 23.37 & 335.7 \\
PPO (D2) & \rev{40.4\sd{2.2}} & \rev{80.66\sd{3.36}} & \rev{74.4\sd{1.8}} & \rev{622\sd{31}} & \rev{10.6\sd{4.7}} & \rev{11.97\sd{0.85}} & -- \\
PPO (D3-B ref) & \rev{38.3\sd{2.0}} & \rev{94.87\sd{1.86}} & \rev{91.4\sd{1.2}} & \rev{742\sd{10}} & \rev{4.1\sd{0.3}} & \rev{3.98\sd{0.52}} & -- \\
PPO (\cadetdata{}) & \rev{38.0\sd{6.2}} & \rev{95.98\sd{4.41}} & \rev{92.0\sd{1.3}} & \rev{698\sd{13}} & \rev{4.8\sd{0.4}} & \rev{3.73\sd{0.60}} & -- \\
PPO (\cadetdata{} w/o TryConnect) & \rev{30.9\sd{4.6}} & \rev{90.54\sd{3.60}} & \rev{91.1\sd{1.1}} & \rev{689\sd{10}} & \rev{4.3\sd{0.5}} & \rev{4.11\sd{0.43}} & -- \\
\ours{} (BC) & \rev{43.1\sd{2.1}} & \rev{78.29\sd{3.05}} & \rev{81.7\sd{1.0}} & \rev{796\sd{80}} & \rev{11.9\sd{2.2}} & \rev{8.38\sd{0.52}} & -- \\
\ours{} @1 & \rev{67.2\sd{1.8}} & \rev{107.22\sd{2.31}} & \rev{91.7\sd{0.6}} & \rev{681\sd{17}} & \rev{10.4\sd{0.6}} & \rev{4.00\sd{0.28}} & \rev{15.5\sd{0.8}} \\
\ours{} @3 & \rev{79.8\sd{0.3}} & \rev{121.30\sd{1.59}} & \rev{95.9\sd{0.3}} & \rev{655\sd{12}} & \rev{9.1\sd{0.6}} & \rev{2.06\sd{0.13}} & \rev{47.6\sd{1.6}} \\
\midrule
\multicolumn{8}{l}{\textit{D3-A (99)}} \\
Reference & 100.0 & 50.61 & 100.0 & 170 & 1.2 & 0.00 & -- \\
FreeRouting @1 & 82.1\sd{1.9} & 46.05\sd{0.52} & 97.7\sd{0.3} & 153\sd{1} & 1.0\sd{0.0} & 0.38\sd{0.05} & 8.6\sd{1.6} \\
KRT @1 & 69.7 & 40.54 & 92.8 & 146 & 3.8 & 1.42 & 1.5 \\
GPT-6 Astra @1 & 83.8 & 45.03 & 93.8 & 138 & 1.8 & 0.97 & 138.9 \\
GPT-6 Luna @1 & 46.5 & 31.30 & 78.5 & 119 & 1.2 & 2.90 & 179.1 \\
PPO (D2) & \rev{78.4\sd{3.3}} & \rev{42.28\sd{0.89}} & \rev{90.6\sd{1.1}} & \rev{185\sd{1}} & \rev{1.9\sd{0.9}} & \rev{1.33\sd{0.19}} & -- \\
PPO (D3-B ref) & \rev{75.6\sd{1.6}} & \rev{42.46\sd{0.40}} & \rev{95.7\sd{0.7}} & \rev{197\sd{4}} & \rev{1.1\sd{0.2}} & \rev{0.58\sd{0.07}} & -- \\
PPO (\cadetdata{}) & \rev{80.2\sd{4.7}} & \rev{44.74\sd{1.19}} & \rev{96.6\sd{0.5}} & \rev{191\sd{1}} & \rev{1.3\sd{0.2}} & \rev{0.45\sd{0.08}} & -- \\
PPO (\cadetdata{} w/o TryConnect) & \rev{76.6\sd{1.1}} & \rev{43.27\sd{0.32}} & \rev{96.3\sd{0.4}} & \rev{192\sd{2}} & \rev{1.1\sd{0.2}} & \rev{0.50\sd{0.07}} & -- \\
\ours{} (BC) & \rev{83.2\sd{0.8}} & \rev{41.61\sd{0.61}} & \rev{94.4\sd{0.4}} & \rev{205\sd{20}} & \rev{2.7\sd{0.5}} & \rev{0.82\sd{0.10}} & -- \\
\ours{} @1 & \rev{89.5\sd{1.0}} & \rev{45.13\sd{0.41}} & \rev{96.6\sd{0.4}} & \rev{187\sd{6}} & \rev{2.4\sd{0.2}} & \rev{0.47\sd{0.09}} & \rev{4.2\sd{0.1}} \\
\ours{} @3 & \rev{96.3\sd{0.3}} & \rev{48.87\sd{0.27}} & \rev{99.2\sd{0.4}} & \rev{171\sd{4}} & \rev{1.7\sd{0.0}} & \rev{0.12\sd{0.06}} & \rev{12.3\sd{0.7}} \\
\bottomrule
\end{tabular}
\end{table}

{\textbf{Fine-tuning ablation.}\quad
\cref{tab:finetune-ablation} gives the results of the fine-tuning variants of \S\ref{sec:ablation}, in which the KL penalty and critic detachment are turned off alone and together. Each variant was fine-tuned from the same BC checkpoints as \ours{}.\par}

\begin{table}[H]
\centering

\caption{\rev{Effect of fine-tuning settings on D3-B test. CP and Rout. are in percent. CP is the mean $\pm$ standard deviation over four training seeds. Vias is the mean via count per board.}}
\label{tab:finetune-ablation}
\small
\setlength{\tabcolsep}{5pt}
\newcommand{\sd}[1]{{\scriptsize$\pm$#1}}
\begin{tabular}{lcc!{\hskip 8pt}rrrr}
\toprule
Method & $\beta$ & Critic detached & CP & $\Delta\Phi$ & Rout. & Vias \\
\midrule
PPO (\cadetdata{}) & -- & no & 38.0\sd{6.2} & 96.0 & 92.0 & 4.8 \\
\midrule
\ours{} & 0.05 & yes & 67.2\sd{1.8} & 107.2 & 91.7 & 10.4 \\
& 0 & yes & 64.0\sd{3.4} & 109.1 & 90.1 & 7.7 \\
& 0.05 & no & 66.5\sd{1.5} & 107.7 & 91.0 & 10.1 \\
& 0 & no & 63.3\sd{2.9} & 109.1 & 88.7 & 6.8 \\
\bottomrule
\end{tabular}
\end{table}

{\textbf{CP confidence intervals.}\quad
\cref{tab:cp-ci} gives the 95\% CIs of CP.
On both test sets, the CI of \ours{} @1 did not overlap that of any PPO baseline.
On D3-B test, it overlapped that of FreeRouting, and the CI of \ours{} @3 did not.
The CIs of the three fine-tuning variants of \cref{tab:finetune-ablation} overlapped that of the base setting and did not overlap that of PPO (\cadetdata{}).\par}

\begin{table}[H]
\centering

\caption{\rev{95\% CIs of CP, from the $t$ distribution over four training seeds for the learned methods and over four runs for FreeRouting. Bottom: the fine-tuning variants of \cref{tab:finetune-ablation} on D3-B test.}}
\label{tab:cp-ci}
\small
\begin{tabular}{lrcrc}
\toprule
& \multicolumn{2}{c}{D3-B test} & \multicolumn{2}{c}{D3-A} \\
\cmidrule(lr){2-3}\cmidrule(lr){4-5}
Method & Mean & 95\% CI & Mean & 95\% CI \\
\midrule
\ours{} @1 & 67.2 & [64.3, 70.1] & 89.5 & [87.9, 91.1] \\
\ours{} @3 & 79.8 & [79.4, 80.3] & 96.3 & [95.8, 96.8] \\
\ours{} (BC) & 43.1 & [39.8, 46.4] & 83.2 & [82.0, 84.5] \\
PPO (\cadetdata{}) & 38.0 & [28.2, 47.8] & 80.2 & [72.7, 87.7] \\
PPO (D2) & 40.4 & [36.9, 43.9] & 78.4 & [73.2, 83.6] \\
PPO (D3-B ref) & 38.3 & [35.1, 41.4] & 75.6 & [73.1, 78.1] \\
PPO (\cadetdata{} w/o TryConnect) & 30.9 & [23.5, 38.2] & 76.6 & [74.9, 78.4] \\
FreeRouting @1 & 65.6 & [59.1, 72.2] & 82.1 & [79.0, 85.1] \\
\bottomrule
\end{tabular}

\vspace{6pt}
\begin{tabular}{lrc}
\toprule
Fine-tuning variant & Mean & 95\% CI \\
\midrule
$\beta=0$ & 64.0 & [58.6, 69.4] \\
Critic detachment off & 66.5 & [64.0, 68.9] \\
Both off & 63.3 & [58.7, 67.8] \\
\bottomrule
\end{tabular}
\end{table}

\section{Training Curves}
\label{app:training-curves}

\begin{figure}[htbp]
\centering
\includegraphics[width=\linewidth]{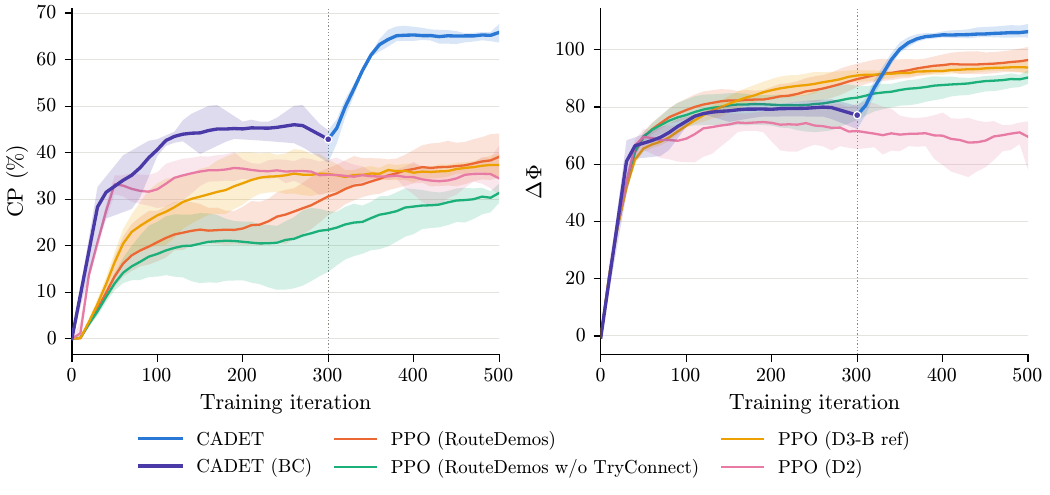}
\caption{D3-B test CP and $\Delta\Phi$ during training, averaged over five rollouts per board. \rev{Lines are means over four training seeds, and bands span the minimum and maximum.} \ours{} (BC) steps are rescaled so that \rev{the steps before fine-tuning} span 300 iterations. The dotted line marks the start of \ours{} fine-tuning.}
\label{fig:training-curves}
\end{figure}

\cref{fig:training-curves} shows D3-B test performance during training.
\rev{Each line is the mean over the four training seeds, and the band spans the minimum and maximum over seeds.}
RL checkpoints were evaluated every ten iterations on all 141 D3-B test boards.
Each value averages five rollouts per board and then averages over boards, so it differs from the @1 protocol of \S\ref{sec:setup}.
D3-B test did not influence checkpoint selection.
RL curves use a centered moving average over five evaluation points.
The \ours{} (BC) curve shows the BC checkpoints saved every 35k steps up to the checkpoint that initializes fine-tuning\rev{, at 280k to 350k steps depending on the seed}.
Its horizontal axis maps \rev{these} BC steps to 300 iterations, and its initial value is the mean random-initialization value of the scratch runs.
\rev{Averaged over seeds, CP was 42.9\% at the end of BC and 65.9\% after 200 fine-tuning iterations.}
No scratch PPO run exceeded \rev{45\%} CP within 500 iterations.

\section{Reproducibility Details}
\label{app:reproducibility}

\cref{tab:reproducibility} lists where the paper specifies each component needed to reproduce \cadetdata{} and the reported results.
Source code for generation, training, and evaluation, together with the D3-B split definitions, is available at {\urlstyle{same}\url{https://github.com/LGAI-Research/CADET}}.
We release \cadetdata{} and the trained checkpoints.

\begin{table}[htbp]
\centering
\caption{Location of the specification for each component.}
\label{tab:reproducibility}
\small
\begin{tabular}{ll}
\toprule
Component & Specification \\
\midrule
Board attribute distributions and fitted values & \S\ref{sec:sampling-distributions}, Appendix~\ref{app:generation-details}, \cref{tab:generator-statistics} \\
Construction and TryConnect search & Algorithms~\ref{alg:generation} and~\ref{alg:connection-search}, Appendix~\ref{app:generation-details} \\
Replay, repair, pruning, and acceptance & \S\ref{sec:generation}, Algorithm~\ref{alg:finalize} \\
Pad-count calibration & Appendix~\ref{app:generation-details}, Equation~\eqref{eq:density-calibration} \\
Generation cost and acceptance rate & Appendix~\ref{app:generation-cost} \\
Data splits and board counts & \S\ref{sec:setup}, Appendix~\ref{app:statistics} \\
Policy architecture and PPO settings & Appendix~\ref{app:training-details}, \cref{tab:ppo-hyperparameters,tab:training-settings} \\
BC settings and checkpoint selection & Appendix~\ref{app:training-details} \\
Evaluation protocol and metrics & \S\ref{sec:setup}, Appendix~\ref{app:metrics} \\
Reward potential & \S\ref{sec:routing-mdp}, Appendix~\ref{app:reward} \\
Software, compute, and rule-based baselines & Appendix~\ref{app:reproducibility} \\
LLM agent settings and prompts & Appendix~\ref{app:llm-prompt} \\
\bottomrule
\end{tabular}
\end{table}

\textbf{Software.}\quad
BC, PPO training, and evaluation all used \kicad{}~9.0.8, Python~3.12, PyTorch~2.8.0 with CUDA~12.8, and Gymnasium~1.2.3.
The rule-based baselines were FreeRouting~v2.1.0 and KRT~v0.15.1.

\textbf{Compute.}\quad
BC took about 97 hours \rev{per seed} on four NVIDIA RTX PRO 6000 GPUs.
Each PPO run used one NVIDIA L40 GPU and 64 environments.
\ours{} fine-tuning took about 85 hours.
Checkpoint evaluation on D3-B ref ran separately on NVIDIA A10 GPUs.
\rev{\cref{tab:compute-cost} compares the cost of \ours{} with that of scratch PPO on D3-B ref and on \cadetdata{}.}

\begin{table}[htbp]
\centering

\caption{\rev{Approximate cost of data generation and training in hours. Data generation is for a single server and includes the independent replay of the demonstrations (Appendix~\ref{app:generation-cost}). The 200 fine-tuning iterations of \ours{} include the critic warm-up. Training times are for a single run.}}
\label{tab:compute-cost}
\small
\setlength{\tabcolsep}{4pt}
\begin{tabular}{llr!{\hskip 8pt}r!{\hskip 8pt}rr}
\toprule
& & & & \multicolumn{2}{c}{Training} \\
\cmidrule(l{0pt}r{0pt}){5-6}
Method & Training data & \makecell[r]{RL\\iterations} & \makecell[r]{Data generation\\(CPU only)} & \makecell[r]{BC\\(RTX PRO 6000 $\times$4)} & \makecell[r]{RL\\(L40 $\times$1)} \\
\midrule
PPO & D3-B ref & 500 & -- & -- & 177 \\
PPO & \cadetdata{} & 500 & 146 & -- & 135 \\
\ours{} & \cadetdata{} & 200 & 146 & 97 & 85 \\
\bottomrule
\end{tabular}
\end{table}

\textbf{Baseline settings.}\quad
FreeRouting and KRT had a time limit of 1,800 seconds per run.
KRT used a 0.1 mm grid, at most three rip-ups, and heuristic weight 1.9, with layers, track widths, clearances, and via sizes read from each board.

\section{LLM Agent Setup}
\label{app:llm-prompt}

Both LLM agents used medium reasoning effort, at most 4,000 output tokens per step, and the last four steps as history.
An LLM episode ended after ten consecutive steps without geometric progress.
The API cost of one run was \$1,131.51 on D3-B test and \$111.21 on D3-A for GPT-6 Astra, and \$6.10 and \$1.54 for GPT-6 Luna.

Each LLM agent step sent a system prompt and a user prompt.
The system prompt (\cref{fig:llm-system-prompt}) holds the task, the routing guidelines, the response format, the S-expression state format, and the static board context.
The user prompt (\cref{fig:llm-user-prompt}) holds the current step, the dynamic board state, the action history, and the valid actions.
\cref{tab:llm-placeholders} describes the placeholders.
The agent parsed its action from the \texttt{<action>} block of each response.

\begin{figure}[p]
\centering
\begin{lstlisting}[style=prompt]
You are a PCB routing agent for KiCad. Connect unconnected pins (points in routing_geometry) within each net. Never connect pins from different nets.

# Priority
1. Complete all connections (reduce unconnected points to zero).
2. Avoid DRC violations (no crossing other nets, respect clearance).
3. Minimize total wirelength.

# Routing Guidelines
- Targets are the (point ...) entries in routing_geometry; the final segment must land on the target pad's (x, y) while the cursor is on that pad's layer (points carry no layer — look up the pad in board_static or the nearby tracks).
- Use exact coordinates from the observation (pad positions, point targets, track endpoints). Only invent coordinates for detour waypoints.
- Avoid static obstacles, board edges, and existing tracks on the same layer (any net, including the current one).
- When a path is blocked, detour: switch layers with `make_via` (offset ≥1mm from any pad — never place a via on a pad), route around on the opposite layer, then switch back.
- If a detour looks too long or also blocked, use PushAndShove mode (`p`) on `make_line` / `make_via` to shove other nets' tracks aside and push the current route straight through.
- When the cursor is already near the remaining target with a clear path, draw the final segment directly with `make_line <x> <y> <mode>` onto the target pad's exact (x, y).
- When the current net's (points ...) list is empty, the net is done — issue `net_end` explicitly to release it before moving on. Continuing with `make_line` / `make_via` on a completed net wastes steps and may trip the no-progress streak.

## Layer-change detour pattern
Route from (0, 0) on layer 1 to (10, 10) on layer 1, bypassing an obstacle via layer 2:
  1. start_route 0.0 0.0 1
  2. make_via 0.0 0.5 w         # offset off the source pad, then cross to layer 2
  3. start_route 0.0 0.5 2      # re-enter routing on the new layer
  4. make_via 10.0 9.5 w        # STOP before the target's (x, y) and cross back
  5. start_route 10.0 9.5 1     # re-enter routing on the target pad's layer
  6. make_line 10.0 10.0 w      # final segment lands on the pad, on its own layer

## Do NOT draw or end on a pad from the wrong layer
Never let `make_line` or `make_via` reach the target pad's exact (x, y) while the cursor is on a different layer than that pad. If you do, you are trapped:
  - placing a via at the pad violates the "no via on pad" rule, and
  - the pad is unreachable from the wrong layer, so the final `make_line` cannot land on it.
Always make the layer-return `make_via` at an offset waypoint (≥1mm away from the target pad) *before* the track arrives at the target's (x, y).

# Response Format
Output exactly one <think>...</think> followed by one <action>...</action>. Never repeat think-action pairs.
"[no effect]" in history = action made no progress; "Last rejected attempt" below = action was refused. Do not repeat either.

# Board State
{state_format_desc}

{board_static}
\end{lstlisting}
\caption{Full system prompt of the LLM agents.}
\label{fig:llm-system-prompt}
\end{figure}

\begin{figure}[p]
\centering
\begin{lstlisting}[style=prompt,title={\small\rev{(a) State format description}}]
The board state is provided in S-expression format. All floating-point values are rounded to 3 decimal places. Each element is enclosed in parentheses: (tag value ...) or (tag (child ...) ...).

## board_static — static board context (unchanged during episode)
  (bbox x y w h)          — board bounding box origin and size in mm
  (scale N)               — max(w, h), for normalization reference
  (net_count N)           — total number of nets
  (copper_layers N)       — number of copper layers
  (boardlines (edge ID (p1 PID x y) (p2 PID x y)) ...)  — board outline edges; an arc outline entry appears as (arc ID (p1 ...) (mid PID x y) (p2 ...)) with mid on the arc between p1 and p2 (full circle: p1 == p2, mid = opposite point)
  (nets (net CODE "NAME" (pads (pad ID x y layer) ...)) ...)   — nets with pad center positions; layer is a copper layer number (1=Top, N=Bottom) or `th` for a through-hole pad (plated through every copper layer — start_route / make_via may land on any layer)
  (obstacles (obs ID x y (size w h) (layer N)) ...)      — static obstacles

## routing_geometry — dynamic per-net routing state (updated every step)
  (net CODE
    (tracks (track ID (p1 PID x y) (p2 PID x y) (layer N)) ...)  — placed track segments
    (vias (via ID (center PID x y) (layers start end)) ...)       — placed vias
    (points (point ID x y) ...))                                  — unconnected ratsnest targets; layer info not provided (infer from the target pad or nearby tracks)

## router_head — current router cursor state
  (xy x y)             — cursor position in mm
  (layer N)            — active copper layer (1=Top)
  (net N)              — currently selected net code (0=none)
  (phase N)            — 0=NET_SELECT, 1=START_ROUTE, 2=ROUTING
  (is_routing true|false)
  (routing_mode M)     — m=MarkObstacles, p=PushAndShove, w=Walkaround
       (MarkObstacles: stop at blockers; Walkaround: detour around them; PushAndShove: shove other nets' blocking tracks aside to force the current route through)
  (step N ratio)       — current step number and step/max_steps ratio
\end{lstlisting}
\vspace{6pt}
\begin{lstlisting}[style=prompt,title={\small\rev{(b) User prompt template}}]
## Step {current_step}
{dynamic_observation}

# History (step {current_step}, {valid_step_count} valid actions taken)
{action_history}{rejected_attempts}

# Valid Actions
Choose exactly one from:
{valid_actions}
\end{lstlisting}
\caption{State format description inserted into the system prompt \rev{(a)} and user prompt template \rev{(b)} of the LLM agents.}
\label{fig:llm-user-prompt}
\end{figure}

\begin{table}[htbp]
\centering
\caption{Placeholders in the LLM agent prompt.}
\label{tab:llm-placeholders}
\small
\begin{tabular}{l>{\raggedright\arraybackslash}p{0.62\linewidth}}
\toprule
Placeholder & Content \\
\midrule
\texttt{\{state\_format\_desc\}} & State format description in \cref{fig:llm-user-prompt}\rev{(a)}, inserted into the system prompt \\
\texttt{\{board\_static\}} & Board outline, nets with pad locations, and obstacles in S-expression form \\
\texttt{\{current\_step\}} & Current step index \\
\texttt{\{dynamic\_observation\}} & Per-net tracks, vias, and unconnected points, and the routing head state \\
\texttt{\{valid\_step\_count\}} & Number of valid actions taken so far \\
\texttt{\{action\_history\}} & Last four steps, each marked ``[no effect]'' if it made no progress \\
\texttt{\{rejected\_attempts\}} & Empty unless the last action was refused, otherwise a ``Last rejected attempt'' block \\
\texttt{\{valid\_actions\}} & Actions allowed in the current routing phase \\
\bottomrule
\end{tabular}
\end{table}

\section{Datasheet for \cadetdata{}}
\label{app:datasheet}

This appendix summarizes \cadetdata{} following the main categories of Datasheets for Datasets~\citep{gebru2021datasheets}.

\textbf{Motivation.}\quad
\cadetdata{} provides engine-level demonstrations for PCB routing. Public board collections hold completed boards, not demonstrations. We used \cadetdata{} to train routing policies by BC (\S\ref{sec:training}).

\textbf{Composition.}\quad
\cadetdata{} contains 100,000 training and 128 validation instances. Each instance is a synthetic two-layer board. It includes the unrouted board and its design rules in \kicad{} format, its demonstration, the routed board that the demonstration produces, and the seed and sampled board attributes. A demonstration is a sequence of engine-level actions, each stored as its type, coordinates, layer, and routing mode. The dataset contains no personal information.

\textbf{Collection process.}\quad
\ours{} generated all instances with D3-B ref as $\mathcal D_{\mathrm{ref}}$ (\S\ref{sec:generation}). Real boards enter only through the attribute distributions and the pad-count calibration. No real-board geometry is copied. Each board is accepted only when fresh-engine replay of its demonstration connects all nets with zero DRVs. Appendix~\ref{app:generation-cost} reports the acceptance rate and runtime.

\textbf{Uses.}\quad
\cadetdata{} is intended for learning PCB routing policies from demonstrations in the engine-level action space. Its attribute distributions are estimated from the two-layer boards of D3-B ref. Each demonstration is one valid routing, not an optimal one \rev{(Appendix~\ref{app:human-comparison})}.

\textbf{Distribution and maintenance.}\quad
\cadetdata{} and its generator are released under the BSD-3-Clause license, as are the synthetic datasets of \pcbworld{}. The D3-B boards come from PCBench~\citep{pcbench2024} and keep the licenses of their source repositories. The release therefore excludes these boards and includes only the statistics that the generator reads. Each release is versioned with the generator and the \kicad{} version used for validation.

\section{\rev{Comparison with Human Layouts}}
\label{app:human-comparison}

{
Each demonstration is a valid routing with a guaranteed clean pass, not an optimal one.
To assess its routing quality, we rerouted the 142 D3-B ref boards from scratch with the procedure that routes the demonstrations and compared the results with the human layouts of these boards.
This procedure, the demonstration search, consists of TryConnect (Algorithm~\ref{alg:connection-search}) and the repair of Algorithm~\ref{alg:finalize}.
The results were scored with the same post-hoc DRC as in \cref{tab:full-results}.
The pad locations of these boards are fixed. Unlike in the generation of \cadetdata{}, the search can therefore leave DRVs or unrouted connections.

\cref{tab:human-comparison} compares wirelength and vias on the 88 boards where the search reached a clean pass.
On these boards, the median per-board ratio of the wirelength of the search to that of the human layout was 1.02, and the search had the shorter wirelength on 44\% of the boards.
The boards that the search completed are likely the easier ones.
\par}

\begin{table}[H]
\centering

\caption{\rev{Demonstration search and human layouts on the 88 D3-B ref boards where the search reached a clean pass. Reference is the original human layout. WL is in millimeters.}}
\label{tab:human-comparison}
\small
\begin{tabular}{lrr}
\toprule
Method & WL & Vias \\
\midrule
Reference & 540 & 3.9 \\
Demonstration search & 553 & 6.0 \\
\bottomrule
\end{tabular}
\end{table}

\end{document}